\documentclass[10pt,twocolumn,letterpaper]{article}

\usepackage{cvpr}              

\usepackage{comment}

\definecolor{cvprblue}{rgb}{0.21,0.49,0.74}
\usepackage[pagebackref,breaklinks,colorlinks,allcolors=cvprblue]{hyperref}

\def\paperID{99999} 
\def\confName{CVPR}
\def\confYear{2026}

\title{Visual General Intelligence: A White Paper}

\author{
Hirokatsu Kataoka$^{1,2}$ \quad
Yoshihiro Fukuhara$^{1,3}$ \quad
Yonglong Tian$^{4}$ \quad
Shangzhe Wu$^{5}$ \quad
Oishi Deb$^{2}$ \quad \\
Ryousuke Yamada$^{1,6}$ \quad
Christian Rupprecht$^{2}$ \quad
Jianyuan Wang$^{2}$ \quad
Kohsuke Ide$^{1,7}$ \quad
Koichi Namekata$^{2}$ \quad \\
Xianzheng Ma$^{2}$ \quad
Yiming Chen$^{2}$ \quad
Robert Geirhos$^{8}$ \quad
Aditi Raghunathan$^{9}$ \quad
Yuki M. Asano$^{6}$ \quad \\
Deva Ramanan$^{9}$ \quad
David Fouhey$^{10}$ \quad 
Andrew J. Davison$^{11}$ \quad
Yilun Du$^{12}$ \quad 
Jiajun Wu$^{13}$ \quad
Zhuang Liu$^{14}$ \quad \vspace{10pt}\\ 
$^{1}$ National Institute of Advanced Industrial Science and Technology (AIST) \quad \\
$^{2}$ Visual Geometry Group (VGG), University of Oxford \quad
$^{3}$ CADDi \quad
$^{4}$ OpenAI \quad \\
$^{5}$ University of Cambridge \quad
$^{6}$ University of Technology Nuremberg \quad
$^{7}$ University of Tsukuba \quad \\
$^{8}$ Google DeepMind \quad
$^{9}$ Carnegie Mellon University \quad
$^{10}$ New York University \quad \\
$^{11}$ Imperial College London \quad
$^{12}$ Harvard University \quad
$^{13}$ Stanford University \quad
$^{14}$ Princeton University \quad 
}

\begin{document}
\maketitle

\begin{abstract}
This paper reconsiders intelligence from a vision-centered perspective and examines whether intelligence emerging from visual experience and learning may provide a pathway toward AGI. In the language domain, beginning with the introduction of the Transformer architecture, the GPT series has demonstrated transfer to unseen tasks through autoregressive language modeling on web-scale text combined with aggressive scaling. This raises a natural question, namely, what capabilities and forms of intelligence can emerge from visual modalities such as images, videos, and geometry? In this paper, we discuss whether visual intelligence can serve as a pathway toward AGI, referred to in this paper as visual general intelligence (VGI), by bringing together contributors from diverse standpoints and affiliations. Our aim is not to offer a single definition of visual intelligence, but to clarify the principles that computer vision should pursue in the AGI era, the visual input modalities, the benchmarks, the learning paradigms, and the relationship between vision, when taken as the core, and other modalities such as language.
\end{abstract}

\section{Introduction}
\label{sec:intro}

Since the early discussions of artificial intelligence (AI), including those surrounding Alan Turing and cybernetics, intelligence has often been discussed as the ability to acquire representations from environmental signals, capture their underlying regularities, generalize to unseen situations, and connect such representations to prediction or action~\cite{turing1950computing,wiener1948cybernetics,legg2007collection,goertzel2014artificial}. At the same time, it is not obvious that intelligence should be reduced to language. 

In the history of Earth, advanced visual systems, including camera-type eyes and compound eyes, date back to the Cambrian period roughly half a billion years ago, and the rapid development of vision has long been discussed as a possible driver of major changes in biological interactions~\cite{parker2003blink,zhao2013complexity}. By contrast, advanced human language, writing systems, and individual languages such as English are all extremely recent phenomena on the timescale of life’s history~\cite{fitch2010evolution,daniels1996world,baugh2013history}. This contrast suggests that vision may be more than a sensor or input modality; the visual modality may instead constitute a foundational intelligent function for capturing the structure of the world.

Recent progress in large language models (LLMs) provides an important reference point for thinking about the emergence of intelligence. Through the lineage of statistical language models, neural probabilistic language models~\cite{bengio2003neural}, distributed representations such as word2vec~\cite{mikolov2013efficient}, sequence-to-sequence models~\cite{sutskever2014sequence}, attention~\cite{bahdanau2015neural}, Transformers~\cite{vaswani2017attention}, BERT~\cite{devlin2019bert}, and GPT~\cite{radford2018improving}, language processing gradually shifted from hand-designed features and task-specific pipelines toward representation learning based on large-scale data and general-purpose learning objectives. GPT-3 was especially striking in that, without images, audio, or embodied interaction, it showed that scaling a simple autoregressive self-supervised objective (next-token prediction) over web-scale text corpora could give rise to broad capabilities such as in-context learning and few-shot adaptation~\cite{brown2020language}. Wei et al. formalized such scale-dependent qualitative capability gains as ``emergent abilities of large language models''~\cite{wei2022emergent}. This also resonates with Sutton's ``The Bitter Lesson''~\cite{sutton2019bitter}: in the long run, methods that scale with computation, data, learning, and search tend to dominate manually engineered knowledge and rules.  Although the success of language models cannot simply be transferred to visual models, the fact that unexpected abilities can arise from a simple objective, large-scale experience, and scaling is a crucial clue for visual intelligence.

Computer vision has also undergone major transformations. Before the era of deep neural networks, the field accumulated rich knowledge based on geometry, statistics, and optimization, including edge and corner detection, local features such as SIFT, stereo, optical flow, structure from motion, and object detection. The success of AlexNet at ILSVRC 2012 marked a turning point, ushering in the deep learning era and leading to dramatic progress in visual recognition~\cite{krizhevsky2012imagenet,russakovsky2015imagenet}. More recently, vision research has expanded beyond supervised learning toward self-supervised learning, image-text contrastive learning, Transformers, image and video generation, 3D/4D reconstruction, robotics, and embodied AI. In this context, a natural question is whether emergent abilities analogous to those observed in language models may also appear when visual models are scaled up and exposed to increasingly diverse visual experiences, or whether vision requires principles specific to visual intelligence.

There is no need to settle this question with a single answer at this stage. One viewpoint is that scaling single-modality, single-objective visual learning to its limit may be sufficient; another is that integrating visual modalities such as video, geometry, and spacetime, or visual tasks such as generation, restoration, and reconstruction, may be essential. What matters is not to make the scaling of vision foundation models (VFMs) an end in itself, but to ask under what learning strategies, data structures, model architectures, and evaluation settings visual representations can generalize to unknown tasks and environments. Current VFMs are increasingly applied to classification, dense prediction, correspondence estimation, and 3D reconstruction, yet there remains substantial room to disentangle whether a capability arises from training data, task design, decoders, prompts, model size, or their combinations. This unresolved nature should be viewed not as a limitation, but as a starting point for asking whether intelligence emerging from vision can evolve into general intelligence.

Recent progress in VLMs and MLLMs is also central to this discussion. CLIP maps images and text into a shared space through contrastive learning, enabling zero-shot recognition mediated by natural language~\cite{radford2021learning}. MLLMs such as Flamingo~\cite{alayrac2022flamingo}, BLIP/BLIP-2~\cite{li2022blip,li2023blip2}, and LLaVA~\cite{liu2023visual} connect image features from visual encoders to LLMs as token sequences or intermediate representations, enabling visual question answering, captioning, dialogue, and instruction following. These models provide powerful frameworks for connecting vision and language and represent one of the central directions for vision research toward the AGI era. At the same time, because image features, image-text pairs, instruction tuning, and reasoning processes are deeply intertwined, it remains difficult to isolate which capabilities originate from visual experience itself and which emerge from the language model. Rather than rejecting MLLMs, this paper keeps multiple viewpoints open, including vision-only, vision-first, language-mediated, and multimodal perspectives, and discusses visual intelligence from several angles.

Visual general intelligence (VGI), as discussed here, is not a rejection of language. Rather, it is a research agenda that asks what vision can understand, predict, and generalize, either before being coupled with language or while interacting with it. Visual data contains dense, observable, and measurable structures. Given such structures, should models predict what will be seen next, generate unobserved regions, reconstruct 3D/4D structures, or integrate multiple objectives to acquire general visual representations? We view visual intelligence research as an attempt to revisit these vision-specific questions through diverse forms of visual learning.

In the AGI era, computer vision has an exceptionally important role to play. Precisely because language models have reshaped the landscape of AI, the computer vision community can now revisit a fundamental question: what kinds of generality can arise from visual learning? The diverse expertise accumulated across visual modalities and tasks offers an unusually rich foundation for exploring VGI. We are entering a creative phase in which the community can proactively design what should be measured beyond benchmark accuracy and incremental improvements, including physical-world applications, adaptation to unknown environments, and societal impact.

VGI, namely intelligence emerging from visual modalities and tasks, is not yet a fully established concept. Precisely for this reason, there is great value in bringing together diverse hypotheses rooted in each researcher’s expertise and experience, and in openly, optimistically, and boldly exploring possible pathways from vision to general intelligence from the perspective of the computer vision community. This paper aims to serve as a starting point for such exploration and to broaden the role of vision research in the AGI era as broadly as possible.\footnote{This paper originates from discussions at the CVPR 2026 VGI Workshop~\cite{vgiworkshop2026}. The views presented here are not intended to provide a single definition of VGI, but rather to summarize multiple perspectives that open diverse possibilities for vision research toward general intelligence.}

\section{Perspectives on Visual General Intelligence}

The path from visual intelligence to general intelligence is unlikely to be captured by a single definition, model, or benchmark. This section brings together diverse research perspectives to help clarify the open questions that the computer vision community may explore.

\subsection{Video models are visual foundation models (Robert Geirhos)}
\label{sec:robert_geirhos}
In the context of my research, visual intelligence can be characterized as the ability to solve any visual task, from perception to reasoning, without being explicitly trained on each new task. Just like large language models are now able to solve most language tasks, I believe that the time has finally come for machine vision to undergo a similar transformation from task-specific to general-purpose models. The computer vision community has built outstanding individual and task-specific models to date \citep[e.g.][]{yang2024depth,kirillov2023segment}, but the field has not yet been unified around a common framework in the same way that LLMs have unified most of natural language processing research. How, then, should we build unified visual foundation models? The foundational primitives that unlocked language intelligence—training large, generative models on web-scale data—apply directly to today's generative video models. I am convinced that transferring these very principles (large-scale training + generative objective) to video models holds the key to solving visual intelligence.

\paragraph{Why \emph{large models} trained on \emph{web-scale data}?} Rich Sutton’s bitter lesson \citep{sutton2019bitter} makes the compelling case that scale has been the single biggest driver of AI progress over the past decades. It is only consequent that LLMs have continued to benefit from larger models and larger, more diverse datasets to date. In a similar vein, there is no reason to believe that video models won't benefit from scale, on both the architecture and data side.

\paragraph{Why \emph{generative} models?} Machine learning models love to learn shortcuts—the easiest solutions they can find to solve a given objective \citep{geirhos2020shortcut}, if only superficially. As a consequence, we need to make the training objective as hard as we can. For example, training classifiers on ImageNet is too easy; they can simply learn to exploit textures \citep{geirhos2019imagenet} or the image background \citep{beery2018recognition} as predictive features, predicting `elephant’ whenever a grey leather-like pattern is detected, or predicting `cow’ whenever grass is present. However, training a \emph{generative} model on visual data requires solving a much harder task. Here, the model cannot simply get away with generating grass without the cow, as the generative objective forces the model to get everything right: the object, the background, the textures, the shapes, the lighting and the shadows---lest it be punished by the training objective. It is this intuition that makes me confident that generative models will be able to solve visual intelligence, simply because their training task is hard enough to be meaningful, forcing them to learn much about the world by virtue of prediction.

\paragraph{And finally, why \emph{video} models?} There is a principled and a practical argument to support the case of video models as visual foundation models. The principled reason is that video generation is the most general framework for many visual tasks, encompassing still images as a special case \citep[e.g.][]{yang2024video}. The practical reason is that in the last two years or so, video models have finally entered the limelight as increasingly capable models, with significant progress happening over the course of months if not days at the moment. In fact, they have become capable enough already to serve as a ``version~1.0'' of visual foundation models: In our recent work~\cite{wiedemer2025video}, we demonstrated that generative video models like Veo~3 exhibit precisely the kind of broad, emergent capabilities that were characteristic of early language foundation models, but in the visual domain instead. Without any task-specific training, the model can perform a wide range of visual tasks simply through image-to-video generation: edge detection, object segmentation, keypoint localization, super-resolution, image editing, style transfer, and even early forms of visual reasoning such as maze solving and graph traversal. Veo~3, a video model trained for entertainment, happens to be a visual foundation model. Subsequent works confirmed and extended these findings in other models too, including open-source models \citep{wang2026very,liu2025can,newman2026video,kim2026collabvr,zhu2026video,zhang2026image,guo2026video,tong2026thinking}. Just like language foundation models quickly progressed from GPT-2 over GPT-3 to, eventually, ChatGPT and modern LLMs, video models are likely to progress towards fully capable visual foundation models as soon as we start building towards this future.\\

\noindent
In light of this, I believe that visual general intelligence is achievable by developing video models as visual foundation models, with the exciting possibility of unifying much of today’s fragmented, ``each task requires its own model'' computer vision community.

\subsection{
Creativity as a Test of Visual General Intelligence (Aditi Raghunathan)}
\label{sec:aditi_raghunathan}

Intelligence is not exhausted by tasks with a single correct answer. In open-ended settings, success also depends on finding creative ways to satisfy a request and making fresh connections that were not explicitly encountered before. For visual intelligence, examples include designing an object under functional constraints, proposing distinct plans for a robot, constructing a scientific diagram, or imagining several ways in which a physical scene could evolve. Creativity in such tasks requires outputs that are not only coherent, but also diverse across repeated generations and original relative to the training experience. These capabilities will become increasingly important as visual models are used for scientific discovery, design, novel training data, and test-time exploration.

Many creative tasks require a leap of thought: an implicit search-and-plan process that coordinates several interdependent decisions before the observable output is generated. In vision, this hidden plan might specify a scene layout, a 3D structure, a set of object relationships, a physical mechanism, a camera trajectory, or a sequence of actions. It may be only indirectly visible in the final image or video. Producing such an output therefore requires capturing higher-order structure across the generation as a whole, rather than assembling independently plausible local choices.

In prior work, we designed minimal algorithmic tasks that are loose abstractions of open-ended real-world tasks and that make creativity measurable in a controlled setting. Drawing on a cognitive-science taxonomy, we considered two forms of creativity. Combinational creativity identifies unfamiliar connections among familiar elements, as in analogy, wordplay, and the discovery of connections between previously separate ideas. Its visual analogue could involve finding an unexpected but meaningful relationship among familiar objects, scenes, concepts, or physical processes. Exploratory creativity constructs new patterns subject to a collection of rules or constraints, as in designing problems, proteins, mechanisms, or narratives. Its visual analogue could be a new scene, 3D structure, video, mechanism, or action plan that satisfies geometric, physical, semantic, or task-level constraints.

In both cases, the goal is to produce solutions that are coherent, distinct, and not simply reproduced from training examples. We refer to this operational notion as algorithmic creativity. A corresponding evaluation of creative visual intelligence should measure at least three properties. First, each output should be coherent, satisfying the relevant physical, geometric, semantic, temporal, or task constraints. Second, generations should be structurally diverse, representing meaningfully different solutions rather than cosmetic changes in texture, color, wording, or viewpoint. Third, outputs should be original relative to the model’s training experience.

Our results suggest that the learning objective matters for acquiring such behavior. On the controlled tasks, multi-token approaches based on teacherless training and diffusion generated more diverse and original solutions than conventional next-token learning. A central issue is global coordination: the latent creative plan is expressed through relationships among several parts of the sequence, and those parts must fit together as a whole. For visual models, this motivates objectives and architectures that can represent a high-level construction and then generate its local details while preserving spatial and temporal consistency.

We also considered where randomness should enter the generation process. Rather than introducing randomness only through output-level temperature sampling, seed-conditioning provides a random input to the model during both training and inference. Different seeds can select different solutions while allowing each solution to remain internally coherent. A visual analogue would use a persistent sample-level latent variable to select a high-level possibility—such as a layout, mechanism, trajectory, or hypothesis—before generating its details. This may encourage diversity among complete plans rather than merely introducing local variation during decoding.

Creativity therefore extends the role of generation in visual general intelligence. A visually intelligent system should not only recognize what is present, reconstruct hidden structure, or predict the most likely future. It should also be able to construct multiple plausible alternatives, preserve the internal logic of each, and search for possibilities that are useful and unfamiliar. Visual models should consequently be evaluated not only by fidelity and accuracy, but also by coherence, structural diversity, originality, and utility. The central question is not simply whether a model can imagine, but whether it can imagine differently without losing structure.

\subsection{A learning system that learns from one datum (Yuki M. Asano)}
\label{sec:yuki_m_asano}

Today’s strongest visual systems are impressive, but they are not general in the way that matters. They are trained once on large, carefully assembled datasets and then deployed as essentially fixed models. They can recognise objects, reconstruct scenes, answer questions about images, or follow textual instructions. Yet they do not build up a coherent understanding of the world through experience, nor do they continue to reorganise that understanding as the world changes.
Today, our field knows one  ``fix'' to this: add more data \& (re-)train.

In my opinion, a \textbf{visually generally intelligent} system should learn from visual experience as it unfolds: continuously, without labels, without repeated reshuffling of the past, and without assuming that the important tasks are known in advance. If we could build such a system, we could have adaptive, continuously improving models and agents that require the equivalent of one light-bulb of energy for updating weights, not that of a city.

The starting point is vision rather than language. Before an intelligent system can describe the world, it must first discover its structure. It must learn which parts of an observation belong together, which objects persist across viewpoints and interruptions, how the observer itself is moving, what changes independently in the scene, and which regularities remain stable over days, months, or years.

The aim is not simply to train a larger \textit{video model}. 
Current video models are powerful and general perceptual modules, but they operate on data that have already been collected, segmented, shuffled, and revisited.\footnote{And I also believe it's a reasonable bet to scale towards a practical definition of VGI with video generative models.} To me, Visual General Intelligence instead asks how a system can learn \emph{during its own visual lifetime}. It should know what it has seen before, detect when something genuinely new has happened, decide which parts of itself should change, and retain knowledge that remains useful. 
This system requires less a continual learning solution, but rather a whole \textit{learning system} that can work autonomously in an open-ended way. Think of a baby: no goal, no talking, no problem. A somewhat rough mental comparison sketch would be today's coding agents that can themselves spawn sub-agents---but instead for vision and from scratch and without language. 

What seems obvious is that this does not get solved by simply having a bigger Transformer and current learning paradigms.\footnote{Happy to be wrong, I like simplicity.} Instead, a plausible artificial visual brain should contain several interacting systems:

\begin{itemize}
    \item \textbf{A perceptual system} that turns the current visual stream into objects, regions, surfaces, motion, and semantic features.\footnote{How does even one best represent video? Frames are redundant, codec is difficult, object-centric somewhat arbitrary.}
    
    \item \textbf{A spatial and dynamic world model} that maintains hypotheses about geometry, viewpoint, object identity, motion, and causal change. It should distinguish changes caused by the observer from changes caused by the world.
    
    \item \textbf{A memory system} that stores information at different timescales.

    \item \textbf{A task and action interface} that exposes the learned visual knowledge to downstream systems, including robots, scientific instruments, search tools, and human users.
\end{itemize}

These modules need not be hand-designed as rigid symbolic components. They may be neural, differentiable, and jointly trained. But they should have distinct roles, update rules, and timescales. The brain is not one homogeneous block, and an artificial learner that must remain stable while continuing to change is unlikely to be one either.

Language enters later as an interface to this visual intelligence. It can name concepts, communicate goals, collaborate, and provide occasional supervision. But language should not be the sole organising principle of the visual system. Much of the world’s structure---depth, motion, persistence, containment, contact, and affordance---exists before it is named. A visually intelligent system should therefore be able to learn these regularities without captions and only subsequently connect them to words. Language becomes a way to query, steer~\cite{steervit}, and teach an already grounded world model, rather than a substitute for one.

The long-term objective is a visual system that grows with experience. It should become more competent, more structured, and more useful the longer it observes the world; without needing to be reset, retrained from scratch, or told in advance everything it will eventually need to understand. 

\subsection{Intelligence will be multimodal, generative, and efficient (Deva Ramanan {\em et al.})}
\label{sec:deva_ramanan}

In this section, I will outline some thoughts on visual intelligence informed by work from our group. I will frame them as ``bets''; some may turn out to be wrong, while some may viewed as obvious. Their intention is to spur (hopefully productive) discussion. Much of these thoughts are motivated by intelligence from an embodied robotics perspective.

{\bf (Visual) intelligence won't be just visual.}  Intelligence will be driven by multimodal signal modeling. Perhaps the elephant in the room is the interplay of vision and language, but other sensory modalities include audio, tactile, and proprioceptive sensing. Such sensory signals are particularly relevant from a robotics perspective. One striking fact is that the vast majority of legged robot demonstrations (including humanoids and quadrupeds) are driven by {\em blind} policies that do not make use of visual observations, but rather proprioceptive feedback that reports the kinematic state of the robot's body. One can move around the world, even acrobatically (via backflips), without opening their eyes. Similarly, there are well-known examples of dexterous manipulation that also operate ``blindly''; one can fan out and identify keys in their pocket via tactile feedback and no vision. In fact, one comment that I've heard in robotic circles is that vision is the {\em least} important perceptual signal for robotics. I would argue against this sentiment and advocate for vision from an information theoretic perspective; compared to other sensory modalities, it is quite high dimensional and tends to contain lots of information about the physical world. This is in fact one reason why it seems attractive to ignore for robotics; it is much easier to build lightweight (and low latency) policies that operate on low-dimensional proprioceptive state. Let me conclude this bet by emphasizing some underexplored research directions. The first is truly multimodal learning. For practical reasons, most multimodal models are really single-modal models that have been post-trained via adaptors for additional modalities. It would be interesting to pre-train representations from raw multi-modal signal streams~\cite{tong2026beyond}. Another direction that I believe is underexplored is tactile perception. While there exists numerous hardware challenges for building scalable tactile sensors, one perspective from which the vision community could immediately engage is through ``near-field'' vision, where visual perception is used to analyze nearby (and possibly touching) scene structures. A wonderful example of this is the GelSight sensor~\cite{yuan2017gelsight}, which reframes tactile sensing as a photometric stereo reconstruction task. Another widely-used example of this are {\em wrist}-mounted egocentric cameras, which could be a rich source for understanding the nearby visual world~\cite{chi2024universal}.

{\bf (Visual) intelligence will be generative.} Here, I subscribe to Hinton's view that ``to recognize shapes, first learn to generate images'' ~\cite{hinton2007recognize}, which itself can be traced back to analysis-by-synthesis roots from pattern recognition~\cite{grenander1996elements,mumford1991computational}. To me, the most shocking and impressive innovation in recent years has been the generative modeling of images and videos. While similar image generation results were arguably already demonstrated with generative adversarial networks (GANs), contemporary diffusion and flow models seem to be a step change, particularly for video. I tend to view generative modeling as the natural evolution and culmination of self-supervised learning. Impressive examples include Veo3 and NanoBanana, pointing to the emergent knowledge that can be gleaned via simple pixel supervision. My bet is that generative models will ``take-over'' representation learning, obviating the need for explicit representation learning in embedded spaces popularized by influential approaches such as DINO and JEPA. Generative models are continually improving with more data and scale, driven by enormous industry investment and support. Indeed, there is a general sense that generative model development requires resources that academia can't provide. Perhaps surprisingly, this also seems to be the case for all but a few players in industry. As such, efficiency is a necessary cornerstone for innovation, which is described further below. That said, I would like to point out some underexplored technical challenges. Open questions include architectures for {\em multimodal generation}; there is a growing body of work on unified models for generating text and images, but given the motivation above, I believe the field should explore generation of multi-{\em sensory} signals. One perspective is that of multi-view learning; while learning from a single modality might be enough to understand the true physical world with enough data and scale, different modalities may produce conditionally independent views of the true world, which can be shown to dramatically improve the efficiency of learning~\cite{blum1998combining}. A second challenge is that of conditioning. Analysis or reconstruction can be viewed as conditional generation; instead of generating an RGBD image, one may generate depth conditioned on RGB. Current approaches for doing so include denoising modalities from with different noise schedules~\cite{duisterhof2026modality}. My hope is that conditional architectures can be made more efficient by learning encoders for the conditioning signal, which itself can be seen as a form of representation learning. 

{\bf (Visual) intelligence will be efficient.} It is striking to point out that one of the first efforts for generative video modeling -- SORA -- was discontinued by OpenAI. The current understanding is that the dollar cost of inference and generation was too high. Serious efforts on video generation seem to be viable only for a few frontier labs, and perhaps out of reach for academia. I believe we need to emphasize the role of efficiency, in both training and inference. 
My bet is that efficiency can be tackled by re-incorporating ``classic'' representations including recurrance, 3D, and multi-scale processing back into generative modeling. One open question is that of long-form video generation - that is, generating videos of 1 minute or longer. It is attractive to apply autoregressive sequence modeling, but a key challenge is maintaining consistent memory. At some level, the underlying requirement is a stateful representation that does not grow over time. This is often phrased as recurrence. For example, when generating a walk-through video of a house, one would expect the living room to look consistent when re-entering after 5 minutes. Here, one can view an online 3D reconstruction as a form of a spatial memory state that enforces consistency. While there appears to be a growing body of work the direction of joint reconstruction and generation, a final underexplored space is the role of resolution. High resolution, high-frame rate video generation is simply too expensive for all but a few. One perspective I've always appreciated is that we should try to model (and generate) the underlying continuous time-world, irrespective of the sensors that are used to capture it. Or put in other terms, perhaps long video generation {\em is
} tenable at low resolution and at low frame rates. Such stop-gap solutions are already common in practical workflows, but I would argue that they should be embraced even further, building off of notable work in the past~\cite{tian2024visual}. I'll finally conclude with a more provocative take on data-driven learning; with the right framework, we don't need that much data. I'll start with the claim that data diversity, and not scale, is the crucial driver for learning. At some level, I think this is obvious; a million copies of the same training example aren't useful for learning. But the reason why scale is so attractive is that it is the {\em easiest} way to assemble a diverse dataset. But with the right curation of data, I posit that scale is not needed. But this adds the equally hard problem of how to correctly (or automatically) curate datasets. I do not have an insightful answer here, but believe the right framing of the problem is that of {\em reinforcement learning} (RL). When compared to other machine learning formulations such as supervised learning or generative modeling, it appears to be the only one that places the responsibility of data collection on the learner. This feels like the right direction! I hope the vision community will embrace it as has robotics and natural language. 

\subsection{Visual Intelligence and Discovery (David Fouhey)}
\label{sec:david_fouhey}

To me, visual intelligence is the ability to understand (and, perhaps, generate) {\it visual} data at the level of a human or better, without an unreasonable amount of supervision on each task. I mean supervision in the broadest sense, not just the direct labels that explicitly trained our past hyper-specialized models (e.g., face detectors), but also the raw data, environments, and simulations that now implicitly train our newer models. 

Currently, I believe that supervision scale is what separates us from visual general intelligence: with sufficient resources, our systems are surprisingly good at many tasks but require too much hand-holding. This usually cues someone to bemoan the situation and contrast AI with the efficiency of a toddler. Our methods are data-hungry, but I think this is the wrong take: more of our problems are solvable with data than we thought, with examples spanning recognition to 3D to generation, and many problems require shockingly small amounts of data to ``brute-force'' to reasonable performance. We just have to figure out how to make things less data hungry. But because of these proof-of-concepts, there is unprecedented demand for visual intelligence, which has opened new opportunities to change the world.

My colleagues will do a fantastic job covering other angles like 3D and video generation, so I will focus on the context of scientific discovery. I have worked in this area for several years, with the goal of making AI-based data that scientists {\it want} to use for discovery. My work includes AI models~\cite{wang2024supersynthia} that reveal the Sun's magnetic field in new detail, as well as AI-powered data~\cite{weeks2025skeletal} that enables reconstructing the process of evolution~\cite{berv2026rates}. I believe this area will only grow, and much of the future visual intelligence budget will be directed at scientific data generated by specialized instruments peering at the many scales of reality. 

I will weave together several problems that I have encountered working with scientists, where there is a mismatch between what vision has to offer and what is needed.

\vspace{1mm}
\par \noindent {\it (Data is limited.)} Computer vision, apart from some frontier lab work, is in the situation where data is a knob one can turn (and indeed likely the most effective knob to turn). However, many key scientific problems are challenging because more data can't be collected. For instance, in understanding the impact of climate change on animal morphology~\cite{weeks2020shared}, one cannot go back and collect species in 1970. Our models applied directly are often poor fits for small amounts of data (and so we need to find ways to align our capabilities to domain needs) and making sense of this data requires integration of other factors.

\vspace{1mm}
\par \noindent {\it (Validation is hard without ground truth.)} Science rarely has the convenient ground-truth we are accustomed to. I mean ground-truth in the broadest sense: not just hand-drawn segments but also other views and data, etc. While some fields are lucky (such as weather with its barometers and rainmeters), most are not. Solar physics has only photons, often from a single angle and time; and evolutionary ecology deals with processes that emerge over the deep time of tens of millions of years. Accordingly, when one makes a new sort of measurement or prediction, there's no leaderboard to compare with and best number to bold in a table. Instead, validation is the long challenging work of reconciling results with our existing knowledge. This shifts the overall fraction of effort much more towards validation than we're accustomed to in common CVPR practice. Despite the best efforts of many, our credit systems as a field are not really prepared for this. 

\vspace{1mm}
\par \noindent {\it (Multimodality, broadly construed, is integral.)} One method of validation is connecting with equations, simplified models, and past data from other instruments. For instance, one can validate estimates of the thermal structure of the Sun's corona~\cite{cheung2015thermal} by connecting them to X-ray emissions observed by another instrument, bridged by the atomic physics. This validation is challenging and requires common sense that comes from other instruments and knowledge (e.g., did we forget a spectral line in the model? is the assumed elemental abundance reasonable? does the model break basic rules?). This sort of connection cannot be automated easily now and primarily consists of work by eye and hand. Unfortunately, the most useful constraints are not the hard rules like Maxell's equations, but instead soft plausibility judgments that, for now, need expert intuition to use. Ultimately, to power scientific discovery, visual intelligence will need to be able to really connect disparate multimodal sources. 

\vspace{1mm}
\par \noindent {\it (Systematics and units can't be ignored.)} Another method of validation is other instruments. In computer vision, we tend to think of data as falling between [0,255] or [0,1], as provided by a black box sensor hardware. In principle, our scientific sensors come with units and so we can indeed use this as a powerful reality check. However, all instruments come with systematics, or non-physical properties. Accordingly, creating and evaluating vision systems for the sciences requires separating out the real signal originating with the underlying phenomenon from the instrument's signals. These require deep understanding to disentangle: the sun oscillates at 11 years~\cite{hathaway2015solar} and 5 minutes, but spurious signals have been found at 160 minutes~\cite{brookes1976observation} and 24 hours~\cite{hoeksema2014helioseismic}. In our experience, models latch onto the real and the fake alike, and emulate them faithfully~\cite{higgins2021fast}. We have to resort to eliminating them at the data source, as one does with weather reanalysis data~\cite{hersbach2020era5}, but future models ought to be able to be more editable to separate out and ignore various signals. As a community, we also largely focus on models and math as the primary contribution, when in reality, clean data can often be as big of a driver of success, if not bigger. 

\vspace{1mm}
\par \noindent {\it (Reality is hard to model.)} One question I often encounter is: why not use simulations? Unfortunately, anything where it's worth taking new data is a place where our understanding of reality is blurry. Accordingly, what models we have are only approximate (or there are multiple competing mechanisms that all explain the data). Making matters worse, we're primarily used to nicely behaved reality. For many problems at the forefront of physics, rendering is itself often an optimization problem, with clean solutions existing only for a limited set of problems. Our existing technology largely does not deal with connecting with complex procedures well, although I hope the advent of effective coding models will make it easier to directly connect our models to models of reality.

\vspace{1mm}
\par \noindent {\it (Vision is only part of the broader picture.)} Pulling back, I have found it humbling to work inside large science projects. In computer vision, we tend to think of the CVPR paper as the world: once you make all of your numbers in table bold, you declare success. But the vision part is often only one part of the pipeline for discovery. When I first started working with scientists, I found that they were interested in largely different things than CVPR reviewers: does my AI data have systematics? are there any mistakes in the AI generated data, and can you catch {\it effectively all} of them? I also found that integrating our results into scientific discovery was challenging: there were typically existing ways of doing things that worked around the limitations of past efforts. Figuring out this interface in order to get data in actual use by scientists has been equally laborious and rewarding.

\subsection{What is the Computational Structure of Spatial AI? (Andrew J. Davison)}
\label{sec:andrew_davison}

In FutureMapping 1 and 2~\cite{Davison:ARXIV2018,Davison:Ortiz:ARXIV2019}, I defined Spatial AI as
the capability for artificial devices to build genuinely rich but
efficient {\bf representations} of their surroundings which enable them to
interact with them in the generally intelligent ways people can.
To quote Pearl, `what humans possessed that other species
lacked was a mental representation, a blue-print of their environment which they
could manipulate at will to imagine alternative hypothetical environments for planning and learning'~\cite{Pearl:TechReport2017}.
I believe that the capability to build high quality, task-independent representations
is the key to the creative visual intelligence that will
enable world-changing products such as an autonomous robot that could
clear and tidy a messy kitchen, or a general lightweight wearable
device that could generally enhance a human's memory and intelligence.
Such products not only place very high demands on their Spatial AI systems,
but also very tight constraints on their embedded implementation
in terms of size, weight, and efficiency. The {\bf computational structure} of these systems --- what is computed and stored where, how everything is connected, and how this relates to likely future developments in computing hardware --- is a crucial and rarely discussed research issue.

\paragraph{Representation}

A representation is a set of values describing the state of a device and the world around it. It must be persistent, but adaptable; always changing as new information is acquired about the world, or as the world itself changes. This enables a robot to predict and plan.
The term ‘world model’ has recently become popular in machine learning
research, but scene representation has a long history in AI and robotics. SLAM (Simultaneous
Localisation and Mapping) research is the
most clear example of how incrementally building persistent world
representations enables consistent, long-horizon behavior. Simple visual SLAM maps of sparse features enable reliable long-term navigation to multiple
waypoints, or guaranteed area coverage.  But the representation possible in real-time visual SLAM
has steadily improved, towards dense maps,
semantic labels, and explicit object-level scene graphs, sometimes with estimated physics and dynamics properties. Additionally, these representations do not necessarily aim to be globally metrically precise, but can be accurate locally and more approximate at large scales.

My argument that visual SLAM research is evolving into Spatial AI is especially because this area has long been driven by live
demonstrations, a focus on efficiency, the use of low-cost hardware, real-time visualizations, and the release of open source code. SLAM researchers have long been obsessed by the key questions of how to continually build a persistent representation from a multi-sensor stream of data without drift and within fixed computation and memory budgets.

\paragraph{Machine Learning and State Estimation}

SLAM is notable within the broader picture of artificial intelligence
in that it has not yet been completely dominated by machine learning. The best
performing methods for localization and sparse mapping are still
mostly hand-designed algorithms using geometry and probabilistic state estimation; and dense and semantic SLAM systems use a
balance, where ML is used to replace part of or add to the
functionality of state estimation (e.g. MASt3R-SLAM \cite{murai2025}).

It is natural to ask whether the whole of SLAM and Spatial AI
will ultimately be performed end-to-end by a neural network, with no
hand-designed representations or algorithms at all, and much recent research on 3D foundation models is moving in this direction. In my opinion, this is not the most important question.
Spatial AI
systems of the future will surely have learned and designed
components so tightly integrated that it is hard to tell which is
which; and this position may be reached either by adding more and more
learned modules to a designed system, or by adding structure to a
purely learned system. 
What remains important in either case, is
the storage and computational structure of the whole system.
If it is to have the properties that real applications demand: efficient, composable, and shareable with other devices and human users; then I believe that either method will converge on similar properties.

\paragraph{Processor Hardware of the Future}

 The move from CPUs to GPUs over the past 15 years as the dominant processors of AI is only the beginning of a trend. The key to efficiency in computation is to fully embrace parallelism, with work done by a very large number of relatively simple cores, but also to reduce data movement around chips and the larger systems they form. This implies a long term move towards processors with very fine-grained parallelism, and with memory distributed across processing tiles, which are connected by sparse communication channels which respect locality. These processors will also inevitably trend towards asynchronous, event-driven operation and low bit-rate or even analog representation of numbers, as these properties also maximize efficiency.

\paragraph{
Mapping Spatial AI Graphs to Hardware
}

If processors and memory do become increasingly granular, with local memory and sparse connections between tiles, then the `bitter lesson' is that algorithms which map best to these designs will perform best. The FutureMapping papers describe how the route forward is to find the graphs in data and processing structures in Spatial AI algorithms, and to as far as possible match these to hardware by essentially laying those graphs down geographically onto chips. Algorithmically, we need to focus on methods which can perform inference and learning via message passing on general, loopy graphs; a strong candidate is belief propagation which has powerful properties of distributed convergence, even with asynchronous or lossy communication.

Elements such as the representation of the scene around a robot have their own integral graph structure, as revealed by many SLAM approaches, and these must be designed to have efficient interfaces to the main real-time processing loops which handle live data from sensors and outputs to actuators. Also, very specialised processors will be located close to (or even within) sensors and actuators themselves, such that the communication between them can be at the abstract level of actionable information rather than large flows of data such as video.

In many ways, we may speculate that the drive towards efficient, modular implementation may drive Spatial AI systems to be increasingly like biological brains, which have huge capability with very high efficiency.

\subsection{Visual Intelligence is Embodied Intelligence (Yilun Du)}
\label{sec:yilun_du}

In this section, I present a perspective on visual intelligence informed by the research of our group. I argue that a central test of visual general intelligence is whether it can enable a robot to perceive, reason about, and act reliably in the physical world. Embodiment is more than a downstream application of visual intelligence: by selecting viewpoints and intervening in the world, an agent can reveal properties that are difficult to infer from passive observation alone, test its predictions, and obtain feedback with which to revise its visual model. Meeting this test requires visual systems that generalize to unfamiliar situations, integrate observations over long temporal horizons into consistent scene representations, connect visual understanding to action, actively acquire informative experience, and continue adapting after deployment. Together, these capabilities form a closed loop in which vision guides action and the consequences of action provide evidence for improving visual understanding.

\paragraph{Building Robust Visual Systems with Generative World Models.}
Embodied agents will routinely encounter observations outside their training distributions, where current visual systems may fail unpredictably---and, in robotic settings, sometimes catastrophically. A central challenge for visual intelligence is therefore to generalize robustly across open-ended changes in viewpoint, environment, object configuration, and interaction.

One promising route toward such robustness is to construct generative world models and formulate perception as inference over the latent scene states that could have produced an observation. Because visual observations arise from recurring physical regularities---including geometry, objects, relations, and dynamics---such models can explain unfamiliar scenes by reusing and recombining previously acquired knowledge. Explicit 3D structure~\citep{li2026structured} and compositional structure~\citep{du2024compositional,wang2025compositional,zhou2024robodreamer} can provide useful inductive biases for data-efficient learning and systematic generalization. This approach follows the long-standing paradigm of ``analysis by synthesis''~\citep{yuille2006vision,kulkarni2015picture}, in which visual understanding is achieved by inferring a latent scene explanation whose synthesized observations agree with the input. Consistent with this view, earlier work from our group showed that generative compositional models can generalize to scenes containing numbers and configurations of objects not encountered during training~\citep{wang2025compositional}.

\paragraph{Constructing Consistent Scene Representations.}
Such inference cannot restart from each individual frame. For a robot to operate autonomously over days or longer, it must integrate a continuous stream of partial observations into a persistent and internally consistent representation of its environment. This representation must associate observations across changes in viewpoint and occlusion, preserve the identities and relationships of entities, and update as the physical world changes. Our group has explored object-centric representations that maintain individual entities over time~\citep{du2022learning,fu2023neuse}, as well as scene-level 3D memories that aggregate observations across an environment~\citep{yang20253d}. A key direction is to develop dynamic, hierarchical representations that connect global scene structure to objects, parts, and their relationships while tracking how each level evolves over time. These representations should support incremental updates and task-dependent retrieval at the appropriate level of abstraction. The goal is therefore not simply a persistent map, but a revisable scene representation that supports prediction, action, and targeted information gathering.

\paragraph{Connecting Visual Intelligence to Action.}
Ultimately, visual intelligence should provide the task-relevant understanding needed for a robot to act, interact, and explore. Visual plans need not reproduce every aspect of an observation with equal fidelity, but they must preserve distinctions that change the actions available to the robot or the outcome of a task. Our group has explored this connection by formulating decision-making as conditional visual generation~\citep{du2023learning,zhen2025tesseract,chen2025large}. In this framework, a generative model proposes a high-level visual trajectory representing how a task could be completed, while a separate low-level controller grounds that trajectory into continuous robot actions. This decomposition allows visual generation to express task-level intent and anticipated physical evolution while delegating precise execution to the controller. Comparing predicted and observed outcomes then turns execution into a test of both the plan and the visual world model.

\paragraph{Active Visual Learning.}
Action is not only an output of visual intelligence; it is also a means of obtaining visual evidence. By moving to informative viewpoints and physically interacting with objects, a robot can resolve perceptual ambiguities and reveal properties, affordances, and dynamics that passive observation may leave uncertain. Changing viewpoint can resolve occlusion, while pushing, lifting, or opening an object can provide evidence about its articulation, stability, and physical behavior. Such interactions can also expose limitations in the robot's current visual model.

Classic evidence from biological vision supports the importance of coupling perception with self-generated action. In Held and Hein's experiment, pairs of kittens received closely matched visual input, but only the kittens whose own movements controlled that input developed normal visually guided behavior~\citep{held1963movement}. Earlier work from our group explored an initial step toward active visual learning by training a curiosity-driven policy to seek regions of an environment that maximize surprisal under a learned visual representation model~\citep{du2021curious}. Such targeted interaction makes action a mechanism for selecting the experience from which the visual system learns.

\paragraph{Continually Adapting Visual Intelligence.}
Active exploration is valuable only if the resulting evidence can be incorporated into the visual model. A robot will inevitably encounter unfamiliar objects, materials, and visual phenomena---for example, transparent or refractive surfaces---that are not adequately represented in its initial training data. Its visual system should therefore remain plastic after deployment, continually revising its representations and world model in light of new experience.

Recent work from our group explores preliminary mechanisms for such self-improvement: SILVR improves a video-based robotic planner using trajectories generated through the agent's own interactions~\citep{luo2026silvr}, while World Action Verifier uses detected prediction failures to guide targeted data collection and world-model refinement~\citep{liu2026wav}. A central challenge is to incorporate genuinely new experience without catastrophically forgetting previously acquired knowledge~\citep{mccloskey1989catastrophic}. One promising direction is a ``wake--sleep'' process that separates online interaction and experience gathering from periodic knowledge consolidation. Together, these components form a closed visual learning loop: infer structured scene explanations, maintain them over time, predict how the visual world will evolve under action, intervene to obtain informative evidence, and revise the model when prediction and observation disagree.

\subsection{Seeing the Physical World via Code (Shangzhe Wu \& Jiajun Wu)}
\label{sec:jiajun_wu}

One way to think about visual intelligence is as the recovery of the structure, or code, of the physical world from visual observations. An image is a measurement of a physical phenomenon: a scene consists of objects with particular shapes and materials, illuminated in a particular way, observed from a particular viewpoint, and moving and interacting under physical laws. The image records only the light that this arrangement happens to send toward the camera. Seeing, on this view, is an attempt to invert that measurement---to infer, from the recorded light, the underlying scene that gave rise to it. The framing is a long-standing one in vision~\cite{marr1982vision,grenander1996elements,mumford1991computational,yuille2006vision}, and it is broadly the one our own research has followed.

The appeal of this framing is that what it recovers remains valid beyond the conditions under which it was observed. Suppose a system has determined that a scene contains a mug of a certain shape and material resting on a table. It is then poised to answer questions, often counterfactual, that were never posed during training: what the mug looks like from behind, what happens if the table is tilted, where a hand should be placed to pick it up. A system that has instead learned the statistics of how mugs on tables tend to appear may well answer some of these correctly, but it has no particular reason to. Structure of this kind is what allows a modest number of observations to support an open-ended range of predictions and actions~\cite{psu2024,phdthesis2019}.

This also shapes how we think about the relationship between vision and language. Language is already an abstraction over experience: it records what people have found worth describing to one another, in the vocabulary they happened to develop. A good deal of how the physical world behaves has never been described in this way---how a particular stroller folds, how dough responds to a rolling pin, how a rose opens and then collapses over the course of a few weeks~\cite{rose4d2025}. Visual observations carry this information regardless of whether anyone has named it, which seems to us a reason to treat vision as a source of world structure in its own right, rather than primarily as an input channel to a language model.

\paragraph{What structure means.}
It helps to be specific about what this structure of the physical world consists of. First, \emph{entities}: a scene decomposes into objects, and objects decompose into parts, recursively. Second, \emph{intrinsics}: the properties belonging to an object itself rather than to the circumstances under which it is observed---its geometry, its texture, its material, how its parts are jointed and how far they can move, and physical properties such as mass, friction, and stiffness~\cite{objectintrinsics2023,magicpony2023,3dfauna2024,anymate2025,particulate2026,li2026instructparticulate,geng2026neurok,galileo2015,rose4d2025}. Third, \emph{extrinsics}: pose, illumination, and camera, which account for a large share of the variation in the pixels while leaving the objects themselves unchanged, and which are therefore worth factoring out. Fourth, \emph{relations}: which object supports which, which part turns which, what is contained in what. Fifth, \emph{dynamics}: how the state of a scene changes when something acts on it~\cite{vda2017,lyu2026choreograph,dynamicssurvey2025}.

This decomposition is by no means exhaustive; it is simply the one that has been productive in our own work. It says what should be recovered, not how it should be encoded: a joint axis may be an explicit parameter or a direction implicit in a network's weights. What is appealing about such a decomposition is that it allows a system to predict precisely the effect of an action: once it represents a joint axis, it can say what happens when the door is pulled; once it represents mass and friction, what happens when the stack is nudged. A representation that leaves these quantities entangled may still reproduce the appearance of a scene very well, while supporting neither.

\paragraph{Structure at scale.}
Progress over the past few years suggests that recovering such structure will require scale, as the visual world is far too varied to be covered by models specified by hand. The more interesting question is what that scale is spent on. Much of it currently goes into generative models trained to predict pixels, which are judged largely by how plausible their outputs look. That criterion is silent on whether shape, material, articulation, or physical properties are represented at all. A video model that renders a convincing falling cup has shown that it knows what falling cups look like; it has not shown that mass, contact, or friction appear anywhere inside it. The gap is hard to detect when evaluation rewards perceptual quality, and becomes evident as soon as one wants to intervene, edit, or act~\cite{worldscore2025,enact2026}.

Whether these quantities eventually emerge from pixel prediction alone, given enough data and compute, remains an open question. Our own work has taken the other route, treating the structure as an explicit target of learning rather than as a by-product of another objective. The physical world is compositional---objects made of parts, scenes made of objects, motions made of joint rotations---so a system that recovers these pieces should be able to reuse them in new combinations, and the hundredth object should cost less than the first. Modelling appearance alone offers less of this, since each new configuration may have to be captured on its own terms. The standard objection to this route is that structured representations do not scale, largely because so much of the structure has had to be put in by hand, and hand-built components do not improve as data and compute grow~\cite{sutton2019bitter}. We would suggest that this is a limitation of how the structure is obtained rather than of structure itself, which puts the weight on how structure is represented and on where it comes from.

\paragraph{Some structure is best written down, and some is not.}
At one extreme, the representation is entirely continuous: the scene becomes a collection of vectors produced by a network. This is expressive and easy to optimize, but no part of it can be explicitly read, edited, or verified. At the other, it is entirely discrete and named: the scene is a mug, resting on a table, with a handle that turns about a given axis. This can be read, edited, and verified, but it reaches only as far as its vocabulary.

An effective representation will need to capture both~\cite{scenelang2025, nscl2019, left2023, hybridworldrep2026}. Structure that is discrete and relational---which parts exist, how many, how they are arranged, what connects to what---is naturally written as a program or in words. Structure that is continuous and hard to name---the exact geometry of a leaf, the precise appearance of a glazed ceramic---is better left as a learned embedding. The Scene Language~\cite{scenelang2025} is one concrete instance: a scene is represented by a program giving its hierarchical structure, words naming the semantic class of each entity, and embeddings capturing the visual identity of each one. The same idea underlies our work on neuro-symbolic concepts, where a concept such as ``left of'' or ``put down'' is stored jointly as a symbolic program and a neural network grounded in perception and action~\cite{nscl2019,left2023,nsconcept2026}.

\paragraph{Coding agents as authors of structure.}
Much of the structure described above can be written as a program. Objects are hierarchical, so programs nest; parts repeat, so programs loop; joints have limits and parts have constraints, so programs take parameters and assertions. LLM agents have recently become excellent at exactly this: they can write a program, run it, read the error, and try again, with enough reliability to be reshaping the software industry. If physical structure can be expressed as code, then systems that are good at writing code may also be good at building models of objects and scenes~\cite{articraft2026,3dgeneralist2026}.

That competence rests on a fundamental property of code itself: it runs and can be validated. A program compiles or it does not, passes its tests or it does not, and this automatic check is what makes coding something an agent can be trained to do well. The property carries over when the program describes an object rather than a piece of software. Such a program can be executed, rendered, and simulated, and its output compared against the image it was meant to explain or the motion the object is meant to have. A proposed structure can therefore be corrected rather than merely produced. Agents already write code that produces articulated 3D assets across hundreds of categories~\cite{articraft2026}, and code that drives robots through manipulation tasks, improving as execution feedback and visual comparison are returned to them~\cite{capx2026}. How far the idea extends is open. Present-day agents lean heavily on the abstractions handed to them and on a language model's prior knowledge of how everyday objects are built and how they work, and they degrade as those priors are withdrawn~\cite{capx2026}. Whether such abstractions can instead be extracted from what is being observed, rather than what is supplied in advance, seems to us the question that matters most.

\subsection{From Visual Models to Vision-Native Intelligence (Zhuang Liu)}
\label{sec:zhuang_liu}

In this section, I do not aim to provide a complete theory of visual general intelligence. Some points may be familiar, and some are speculative. I include them because they reflect a few beliefs shaped by my own research taste: simple scalable methods, empirical understanding of models, and the role of data and evaluation. Even as visual models become stronger, many AI systems still use vision mostly as an attachment to language. I would like to think about VGI in a more vision-native way, where vision is a natural channel for acquiring context and providing timely feedback.

\paragraph{Why vision looks less central today.}
Human vision is an extremely strong baseline. It has been shaped by a long evolutionary process, and most people use it with little conscious effort. We recognize objects, navigate cluttered environments, and correct our actions through visual feedback almost continuously. This may partly explain why language progress looks so striking. In language, AI systems compete against a more variable and culturally trained human baseline. In vision, they compete against one of biology's most mature intelligence systems.

This does not make vision less important. It suggests the opposite. Vision is one of the main ways humans experience and organize the world. Human thought is often expressed in language, but many internal cognitive processes are visual or spatial. We imagine scenes, reason about object relations, and plan actions in ways that are not naturally text-token-like. I do not know whether this is the missing ingredient in current AI systems. But I suspect that text-dominated training may miss an important kind of experience: grounded, spatial, continuous, and action-oriented visual experience.

\paragraph{Why scaling vision is different from scaling language.}
A central difficulty is that vision does not come with symbolic units in the same way language does. Language has a relatively clear hierarchy: words, phrases, sentences, and documents. Modern language models use subword tokens, but these tokens are still built on top of a symbolic system that humans already created. Language is therefore not raw experience; it is a compressed form of human experience. Vision is different. Pixels are too low-level to be units of intelligence. Patches are engineering conveniences rather than perceptual primitives. Human visual understanding seems object-centric and hierarchical, involving scenes, objects, parts, and relations~\cite{marr1982vision,gibson1979ecological}. Yet there is no agreed-upon hierarchy for vision comparable to the hierarchy of language.

This makes visual learning especially expensive. A short piece of text can summarize what happened in a room, but the corresponding visual stream contains far more information: appearance, motion, geometry, lighting, and local details that may or may not matter. Vision has no equivalent compression layer. Without useful intermediate units, visual learning must discover what to keep, what to ignore, and how to compress the input on its own. Earlier vision systems imposed tasks such as detection, segmentation, or depth estimation. These abstractions are useful, but they are not equivalent to language tokens. A bounding box or a mask is not comparable to a word or a sentence in expressive flexibility. I therefore do not think the answer is to manually design a fixed visual vocabulary. Some forms of visual intelligence may require several orders of magnitude more effective compute and data processing than language pre-training, perhaps 1,000x or even 10,000x in some settings, before the right abstractions emerge naturally.

The visual world can be decomposed into arbitrarily fine details, but not all details are useful. Representing everything at the finest possible granularity would waste enormous compute. Representing everything too coarsely would miss what matters. The useful unit of visual perception is therefore task-dependent. A visual system should keep coarse representations for irrelevant regions and allocate fine-grained representations to task-relevant regions. In this sense, the problem of visual units is inseparable from the problem of where to look. This view is consistent with active vision and visual routines, where perception is an active process of selecting and organizing information for behavior~\cite{yarbus1967eye,ullman1984visual,ballard1991animate}.

This is not an argument against scaling. In fact, I am a believer in scaling. In my own research experience, the simplest approaches often work best when they have the right data, the right objective, and enough scale. I therefore do not want to argue that we should hand-design visual units, attention policies, or active perception modules. With enough compute, sufficiently large models, and sufficiently diverse visual experience, many aspects of adaptive visual abstraction and where-to-look behavior may emerge without explicit engineering. The question is not scale versus structure, but what kind of scale is needed for the right structures to emerge, and how to make that scaling efficient enough to be practical.

\paragraph{What this suggests for VGI.}
Raw scale is not the same as visual diversity. More images or videos do not necessarily produce broader visual experience if the data is repetitive, narrow, or biased. Large-scale visual datasets are not automatically diverse. Even datasets that appear broad can contain strong dataset-specific signatures, and understanding these biases is part of understanding what visual experience our models actually receive~\cite{liu2024decade,zeng2024understanding}. Besides improving model skill and scaling compute, we should better understand what kinds of visual experience our datasets provide, what they miss, and how to curate data that is broader, less biased, and more useful for learning general visual abstractions.

Most current systems are not designed around vision as a continuous or low-friction channel of context. They require users to manually capture visual information, upload it, and explain why it matters. A more vision-native system would invert this relationship. It would treat visual context as part of the natural interaction between the system, the user, and the world. This becomes especially important as AI systems become more autonomous and world-facing. A robot, a laboratory assistant, or a medical system cannot rely on humans to continuously describe the state of the world~\cite{driess2023palme}. It must perceive for itself, decide what matters, detect changes, and use feedback to correct its actions. Vision is not behind language in importance; it is behind language in current interface visibility.

I hope future work on VGI will move beyond static image or video question answering. We should ask not only what a model can infer from a given visual input, but whether it knows what it needs to see next. We should also evaluate adaptive visual abstraction: the ability to keep irrelevant regions coarse while inspecting task-relevant details. For me, the key test is simple: can a visual system learn when to look, where to look, and how much detail it needs?

\subsection{Visual intelligence towards general intelligence (Hirokatsu Kataoka~\emph{et al.})}
\label{sec:organizer}

We view visual intelligence as the capacity to acquire usable knowledge of the world from visual experience and apply it to prediction, imagination, reconstruction, and new tasks. In this sense, visual intelligence is intelligence that emerges from visual experience: the ability to learn usable structure of the observable world from partial visual observations, understand its hidden structure, predict what may happen next, reconstruct what is not directly observed, imagine possible alternatives, and generalise such knowledge to new visual tasks. It is not simply a stronger image classifier, nor simply a visual encoder connected to an LLM.

One important consequence of this view is that human vision should not be the upper bound of visual intelligence. A system exposed to richer-than-human visual information may acquire intelligence that goes beyond human-centered vision. The goal of visual intelligence is not to imitate human perception, but to ask what kind of intelligence can emerge when machines learn from visual experience at a scale, diversity, and precision that humans cannot access. This view also suggests a common analogy across humans, animals, and machines. Biological organisms can exhibit intelligent behavior without language by sensing the world, discovering regularities, predicting, and acting. Vision and perception existed long before symbolic language, and it is plausible that intelligence-like functions can emerge without language. The critical question for vision research is whether a sufficiently powerful visual system can acquire internal knowledge of the world and elicit intelligent behaviour from it.

If a model understands the world deeply enough, many abilities may already be latent inside the model. The remaining problem may be how to elicit them. This is one of the most important lessons from the development of language models: a model trained with a simple autoregressive objective over large-scale text could perform new tasks when given only a few examples in context. The visual analogue may be true visual in-context learning. A model would be shown a few visual examples of a new task, infer the task structure, infer the appropriate output format, and solve new cases without explicit retraining. In other words, if a model captures the appearance and dynamics of the world sufficiently well, visual in-context learning may make it easier to elicit new capabilities without parameter updates. Even then, a world model is necessary but not sufficient. Intelligence also requires the ability to use that model under new goals: to infer tasks, choose relevant abstractions, generate counterfactuals, adapt to new observations, and act or respond appropriately.

Our central hypothesis is that visual intelligence may emerge through the convergence of three complementary learning objectives: sequential prediction, open-ended generation, and reconstruction. These correspond to three fundamental operations of intelligence. Sequential prediction anticipates what comes next. Open-ended generation explores what could exist. Reconstruction infers hidden structure from incomplete observations. Individually, each may produce powerful models. Together, they may define a path from vision foundation models to visual intelligence, and eventually to general intelligence.

Autoregressive visual learning may become the visual counterpart of GPT. In language, next-token prediction appears simple, but successful prediction requires a model to capture grammar, semantics, facts, discourse, reasoning patterns, and task structure. In vision, predicting the next visual state may require an even richer set of latent concepts. Given an image, video, or embodied observation, the model must infer objects, geometry, motion, occlusion, physical constraints, camera movement, agent behaviour, temporal causality, and uncertainty. To predict coherently is not simply to extrapolate pixels; it is to infer the hidden structure that makes the next observation plausible. If a model can predict future visual states across diverse environments and long horizons, it is being forced to learn how the world evolves. This pressure is a natural route toward visual world modeling: the model must learn from observations, infer their structure, and anticipate what should follow.

Generative learning is learning to imagine. Generation can impose broader modelling demands than recognition, because recognition often succeeds by detecting discriminative cues, whereas generation requires a model to represent what could exist. A high-quality generative model must learn appearance, geometry, material, lighting, motion, interaction, and plausibility. If such a model can generate high-fidelity images, videos, 3D scenes, and possible futures of the physical world, it becomes more than a content generator. It may provide a basis for an internal simulator, provided that it captures physical and temporal structure rather than appearance alone. A visual agent that can imagine can continue learning inside its own generated possibilities: it can create counterfactuals, test hypotheses, simulate consequences, and refine its understanding before acting. In this sense, imagination is not decoration. It is a mechanism for learning, understanding, and thinking.

Reconstruction provides a direct route to learning hidden structure from incomplete observations. Super-resolution, image restoration, inverse rendering, depth estimation, multi-view reconstruction, 3D reconstruction, and 4D dynamic reconstruction all require the model to recover hidden causes from incomplete observations. Among these, scalable 3D reconstruction is especially important. A system that can reconstruct the world accurately from images on the fly must acquire notions of space, scale, correspondence, surfaces, objects, motion, and physical consistency. Such a system may also learn to detect anomalies, infer invisible structure, and understand why some observations are physically plausible while others are not. If reconstruction becomes dramatically stronger, it may no longer be merely a recovery task. It may become a way of recovering regularities of the physical world from visual evidence.

The role of visual intelligence in the journey toward AGI is therefore not to replace language, but to provide a grounded and potentially independent path. Language-first AI has demonstrated that broad capabilities can emerge from simple objectives at scale. Vision-first AI should now ask the analogous question: what emerges when models learn from visual experience at scale? The early signs are already visible in systems related to video understanding, geometry understanding, physical understanding, embodied intelligence, language alignment, and egocentric perception. These should not be viewed as isolated demos, but as early coordinates of a larger map from visual foundation models to visual intelligence and from visual intelligence to general intelligence.

\section{Summary and Discussion}

\subsection{Summary of Perspectives}

The perspectives presented in this paper approach visual general intelligence (VGI) from complementary directions. Table~\ref{tab:vgi_positions} provides a concise summary of their central positions.

\begin{table*}[t]
\centering
\small
\renewcommand{\arraystretch}{1.15}
\begin{tabular}{p{0.20\textwidth} p{0.74\textwidth}}
\hline
\textbf{Contributor} & \textbf{Position} \\
\hline
Robert Geirhos &
Generative video models will become visual foundation models. \\

Aditi Raghunathan &
Creativity, measured through coherence, structural diversity,
originality, and utility, should be a central test of visual general
intelligence. \\

Yuki M. Asano &
VGI should continuously and autonomously learn from visual experience,
rather than remain fixed after large-scale pre-training. \\

Deva Ramanan et al. &
Visual intelligence should be multimodal, generative, and
computationally efficient, combining vision with signals such as touch,
audio, and proprioception. \\

David Fouhey &
Visual intelligence should support scientific discovery from specialized and limited observations through validation against physical knowledge, other instruments, and sensor systematics. \\

Andrew J. Davison &
Spatial AI requires persistent world representations that integrate
state estimation, learning, real-time computation, and efficient
hardware. \\

Yilun Du &
Visual intelligence is embodied intelligence, connecting perception,
generative world models, action, active exploration, and continual
adaptation. \\

Shangzhe Wu \& Jiajun Wu &
Visual intelligence should recover compositional physical structure
from visual observations, so that its representations can support
editing, simulation, verification, and action. \\

Zhuang Liu &
VGI should be vision-native, learning when to look, where to look, and
how much visual detail is required for the current task. \\

Hirokatsu Kataoka et al. &
Prediction, imagination, and structural understanding (realized
through autoregressive, generative, and reconstruction learning) may
provide a path from visual experience to general intelligence. \\
\hline
\end{tabular}
\caption{A concise summary of the perspectives on visual general
intelligence presented in this paper.}
\label{tab:vgi_positions}
\end{table*}

Taken as a whole, the positions can be broadly grouped into the following directions: scaling and generative modeling; creativity, prediction, imagination, and reconstruction; continual and lifelong learning; persistent 3D and spatial world representations; embodiment, active perception, and action; multimodal learning and computational efficiency; scientific discovery and generalization to unfamiliar environments; the recovery of compositional physical structure; and forms of visual intelligence that are not organized primarily around language.

Several contributions assign a central role to generative learning. Geirhos argues that generative video models trained on web-scale visual data can acquire general-purpose capabilities because generation requires a model to account for substantially more of the visual world than task-specific recognition (Section~\ref{sec:robert_geirhos}). Raghunathan extends generation beyond fidelity toward creativity, requiring coherent, structurally diverse, original, and useful outputs (Section~\ref{sec:aditi_raghunathan}). Ramanan et al.\ view generation as a possible culmination of self-supervised representation learning, while emphasizing multimodality and efficiency (Section~\ref{sec:deva_ramanan}). Du uses generative world models for robust perception and visual generation as an interface to action (Section~\ref{sec:yilun_du}). Kataoka et al.\ treat generation as imagination and combine it with autoregressive prediction and reconstruction-based structural understanding (Section~\ref{sec:organizer}).

Shangzhe Wu \& Jiajun Wu provide a complementary qualification to this emphasis on generation (Section~\ref{sec:jiajun_wu}). They argue that predicting pixels or producing photorealistic outputs does not by itself demonstrate an understanding of the physical world. Visual intelligence should recover the entities, intrinsic and extrinsic properties, relations, and dynamics that generated an observation, so that the resulting representation can support interventions, simulation, verification, and action.

Other contributions emphasize how visual knowledge should be acquired, maintained, and used over time. Asano argues that a visually intelligent system should develop throughout its visual lifetime, rather than remain fixed after pre-training (Section~\ref{sec:yuki_m_asano}). Davison emphasizes persistent but adaptable spatial representations, closed-loop state estimation, and the computational structure needed for real-time Spatial AI (Section~\ref{sec:andrew_davison}). Du connects persistent scene representations to action, active exploration, and continual adaptation. Liu further argues that a vision-native system must determine when to look, where to look, and how much visual detail is needed, rather than passively processing a predetermined visual input (Section~\ref{sec:zhuang_liu}).

The scope of visual intelligence also extends beyond ordinary visual recognition. Ramanan et al.\ emphasize raw multimodal experience, including touch, audio, and proprioception. Fouhey’s scientific setting emphasizes validation without conventional ground truth, the handling of sensor systematics and units, and integration with domain knowledge (Section~\ref{sec:david_fouhey}). Raghunathan's account of creativity similarly asks whether a system can construct possibilities that were not explicitly present in its training experience.

These perspectives do not converge on a single model, learning objective, representation, or definition of VGI. They nevertheless share an important premise: visual general intelligence should not be understood simply as a more accurate image recognition system or as a visual encoder attached to a language model. Instead, VGI concerns intelligence that emerges from visual experience and enables a system to discover and verify the structure of the world, maintain and update knowledge over time, predict possible futures, imagine alternatives, reconstruct unobserved structure, and adapt its behavior to unfamiliar situations. Such knowledge should ultimately be tested not only by whether a system can describe or reproduce an observation, but also by whether its representation can be edited, simulated, and used to support reliable action.

The resulting research agenda therefore extends beyond model scale alone. A central challenge is to determine how generation, compositional physical structure, memory, spatial understanding, embodiment, active information acquisition, multimodal experience, and continual learning should be combined within a general visual system. The diversity of positions in this paper suggests that these components should not be treated as isolated capabilities, but as complementary parts of a broader pathway from visual foundation models to visual intelligence and, potentially, to general intelligence.

\subsection{Discussion}

\paragraph{Can intelligence emerge from visual experience?}

Before asking whether visual intelligence can lead to general
intelligence, a more immediate question is whether intelligence can
emerge from visual experience itself. Current AI systems often treat
vision as an input while assigning abstraction and reasoning primarily
to language. Several perspectives in this paper challenge this
division. Asano argues that objects, depth, motion, persistence, and
affordance can be learned before they are named. Liu similarly
emphasizes that grounded, spatial, and action-oriented experience is
not naturally reducible to text tokens. Shangzhe Wu \& Jiajun Wu add
that visual observations contain physical properties and dynamics that
may never have been described in language. Together with Kataoka
et al., these positions leave open a grounded and potentially
independent path from visual experience to intelligence.

This does not make language or other modalities unnecessary. Ramanan
emphasizes audio, touch, and proprioception, while Du connects vision
to action and physical feedback. The open question is whether visual
grounding should precede multimodal integration or emerge jointly with
it. In either case, vision should not be reduced to an input channel
for an otherwise language-centered system.

\paragraph{Scaling, generation, and physical structure.}

Geirhos argues that generative models trained on large-scale
video data may acquire generalized capabilities across a range of  downstream tasks that are not specified
individually during training. Other positions similarly regard these large-scale generative models as a major milestone in different contexts: Ramanan connects them to representation learning, Du to
world modeling and action, 
Raghunathan to creativity, and Kataoka
et al.\ to imagination.

However, scaling the volume of training data alone is not  sufficient. 
Liu and Ramanan emphasize the importance of 
collecting diverse visual data, rather than simply exposing models to vast quantities of
data in similar patterns. Meanwhile,  Fouhey highlights the need for techniques capable of capturing scientific phenomena only available in a small amount of data. 
Shangzhe Wu \& Jiajun Wu further distinguish photorealistic appearance from physical
understanding. 
A model may generate a visually plausible  event without
understanding the underlying physical properties that govern it. We may therefore need
to investigate representations of physical structure that
can be inspected, edited, simulated, and verified.

\paragraph{Continual learning and evaluation.}

A system trained once and then frozen cannot accommodate an open and
changing world. Asano defines VGI as a system that develops throughout
its visual lifetime, while Du argues that an embodied system must
collect informative experience and continue adapting after deployment.
Davison's persistent spatial representations and Ramanan's stateful
long-horizon models highlight the need for compact memory that can be
accessed and revised. Liu and Du further emphasize that an intelligent
system should actively decide what to observe and use action to resolve
uncertainty.

These requirements also change how VGI should be evaluated. Fouhey’s scientific setting emphasizes validation without conventional ground truth and the handling of sensor systematics and units;
Raghunathan adds coherence, diversity, and originality; Davison and Du
require persistent spatial understanding and reliable action; and
Shangzhe Wu \& Jiajun Wu ask whether recovered structure supports
intervention and simulation rather than reconstruction alone. VGI
benchmarks should therefore evaluate transfer, continual adaptation,
active observation, persistent world knowledge, creativity, physical
consistency, and efficiency. No single benchmark is likely to measure
all of these dimensions, but isolated static tasks are unlikely to be
sufficient.

\paragraph{From visual functions to visual intelligence.}

Computer vision has progressed by decomposing visual understanding
into tasks such as recognition, reconstruction, generation, mapping,
and action prediction. This remains useful, but it also encourages
each task to be addressed by a separately trained system. A general
visual system would instead need to operate across tasks, timescales,
environments, representations, and potentially sensory modalities. It
would preserve and revise knowledge, determine which observations
matter, and connect perception to prediction and action.

If vision foundation models acquire these properties, recognition,
reconstruction, generation, spatial mapping, and action may become
different ways of accessing and testing knowledge held by the same
system. The transition from visual foundation models to visual
intelligence would therefore mark a new phase of computer vision:
from constructing increasingly capable visual functions to
investigating how intelligence itself may emerge from visual
experience.

\section{Conclusion}

This paper has examined visual general intelligence as a research agenda concerned with two related questions: how intelligence may emerge from visual experience, and how such visual intelligence may provide a path toward general intelligence. The perspectives collected here do not support a single definition, architecture, learning objective, or route to VGI. Rather than forcing these positions into one unified forecast, this paper presents several plausible and complementary conclusions. Visual intelligence may emerge through large-scale generative learning, prediction, reconstruction, persistent spatial memory, continual learning, embodiment, active perception, multimodal experience, or combinations of these mechanisms that have not yet been fully explored.

This plurality is itself one of the central conclusions of the paper. VGI should not be defined prematurely as a particular model or capability. Nor should it be reduced to a stronger image classifier, a video generator, a world model, or a visual encoder attached to a language model. The positions presented here instead describe visual intelligence as a broader capacity to acquire knowledge from visual experience, discover and maintain the structure of the world, imagine and evaluate possible alternatives, adapt to new observations, and use visual knowledge for prediction, discovery, and action.

Existing systems demonstrate important components of visual intelligence, but not yet their integration into a general, continually adaptable visual system. Current models nevertheless provide partial and increasingly suggestive evidence. Generative video models exhibit unexpected cross-task capabilities; reconstruction models recover increasingly rich geometric and physical structure; spatial systems maintain persistent representations over time; creative models construct multiple plausible alternatives; and embodied agents improve their perception and behavior through interaction. These capabilities remain fragmented, computationally expensive, limited in temporal extent, or dependent on carefully selected data and training conditions. However, taken together, they may represent early coordinates of a broader transition from visual functions to visual intelligence.

It may therefore be premature to ask whether VGI has already become a path to general intelligence. The more immediate milestone is to determine whether intelligence can first be observed to emerge from visual experience itself. Evidence for this transition should extend beyond improved performance on a fixed collection of visual tasks. It should include transferable capabilities, persistent and revisable world knowledge, creative and counterfactual reasoning, active information seeking, continual adaptation, spatial and physical consistency, and reliable action in unfamiliar situations.

Computer vision may now be entering a new phase in which its central object of study expands from individual visual functions to the emergence of visual intelligence. The purpose of this paper is not to declare which proposed path will prevail, but to make their assumptions, relationships, and open questions visible. The decisive principle may come from one of the directions discussed here, from their convergence, or from an approach that has not yet been imagined. The question left to the field is therefore not only how to build more capable visual models, but what evidence would convince us that intelligence has begun to arise from vision -- and how that intelligence might eventually contribute to general intelligence.

\section*{Acknowledgments}

This work was conducted by AIST researchers under FRONTia, a Japanese national program led by the Ministry of Economy, Trade and Industry of Japan (METI) and the New Energy and Industrial Technology Development Organization (NEDO), aimed at developing a domestic multimodal foundation model for AI robots and physical AI. This work was supported by Japan Science and Technology Agency (JST) as part of Adopting Sustainable Partnerships for Innovative Research Ecosystem (ASPIRE), Grant Number JPMJAP2518.
We would also like to thank the invited speakers of the CVPR 2026 Workshop on Visual General Intelligence (VGI) for their insightful talks and discussions that helped motivate this white paper. In particular, we are grateful to Matt Deitke, Alexei A. Efros, Kristen Grauman, Alex Kendall, and Andrea Vedaldi, who contributed to the workshop. Their perspectives helped clarify the diversity of open questions surrounding visual intelligence.

{
    \small
    \bibliographystyle{ieeenat_fullname}
    \bibliography{main}

@String(CVPR= {IEEE Conf. Comput. Vis. Pattern Recog.})

@String(ICCV= {Int. Conf. Comput. Vis.})

@String(ECCV= {Eur. Conf. Comput. Vis.})

@String(ICLR = {Int. Conf. Learn. Represent.})

@String(AAAI = {AAAI})

@String(CVPR  = {CVPR})

@String(ICCV  = {ICCV})

@String(ECCV  = {ECCV})

@String(ICLR  = {ICLR})

@article{tian2024visual,
  title={Visual autoregressive modeling: Scalable image generation via next-scale prediction},
  author={Tian, Keyu and Jiang, Yi and Yuan, Zehuan and Peng, Bingyue and Wang, Liwei},
  journal={Advances in neural information processing systems},
  volume={37},
  pages={84839--84865},
  year={2024}
}

@article{duisterhof2026modality,
  title={Modality Forcing for Scalable Spatial Generation},
  author={Duisterhof, Bardienus Pieter and Ramanan, Deva and Ichnowski, Jeffrey and Johnson, Justin and Park, Keunhong},
  journal={arXiv preprint arXiv:2606.13676},
  year={2026}
}

@inproceedings{blum1998combining,
  title={Combining labeled and unlabeled data with co-training},
  author={Blum, Avrim and Mitchell, Tom},
  booktitle={Proceedings of the eleventh annual conference on Computational learning theory},
  pages={92--100},
  year={1998}
}

@article{mumford1991computational,
  title={On the computational architecture of the neocortex: I. The role of the thalamo-cortical loop},
  author={Mumford, David},
  journal={Biological cybernetics},
  volume={65},
  number={2},
  pages={135--145},
  year={1991},
  publisher={Springer}
}

@book{grenander1996elements,
  title={Elements of pattern theory},
  author={Grenander, Ulf},
  year={1996},
  publisher={JHU Press}
}

@article{hinton2007recognize,
  title={To recognize shapes, first learn to generate images},
  author={Hinton, Geoffrey E},
  journal={Progress in brain research},
  volume={165},
  pages={535--547},
  year={2007},
  publisher={Elsevier}
}

@article{chi2024universal,
  title={Universal manipulation interface: In-the-wild robot teaching without in-the-wild robots},
  author={Chi, Cheng and Xu, Zhenjia and Pan, Chuer and Cousineau, Eric and Burchfiel, Benjamin and Feng, Siyuan and Tedrake, Russ and Song, Shuran},
  journal={arXiv preprint arXiv:2402.10329},
  year={2024}
}

@article{yuan2017gelsight,
  title={Gelsight: High-resolution robot tactile sensors for estimating geometry and force},
  author={Yuan, Wenzhen and Dong, Siyuan and Adelson, Edward H},
  journal={Sensors},
  volume={17},
  number={12},
  pages={2762},
  year={2017},
  publisher={MDPI}
}

@article{tong2026beyond,
  title={Beyond language modeling: An exploration of multimodal pretraining},
  author={Tong, Shengbang and Fan, David and Nguyen, John and Brown, Ellis and Zhou, Gaoyue and Qian, Shengyi and Zheng, Boyang and Vallaeys, Th{\'e}ophane and Han, Junlin and Fergus, Rob and others},
  journal={arXiv preprint arXiv:2603.03276},
  year={2026}
}

@misc{legg2007collection,
  title        = {A Collection of Definitions of Intelligence},
  author       = {Legg, Shane and Hutter, Marcus},
  year         = {2007},
  eprint       = {0706.3639},
  archivePrefix= {arXiv},
  primaryClass = {cs.AI}
}

@article{goertzel2014artificial,
  title   = {Artificial General Intelligence: Concept, State of the Art, and Future Prospects},
  author  = {Goertzel, Ben},
  journal = {Journal of Artificial General Intelligence},
  volume  = {5},
  number  = {1},
  pages   = {1--48},
  year    = {2014}
}

@article{turing1950computing,
  title={Computing Machinery and Intelligence},
  author={Turing, Alan M.},
  journal={Mind},
  volume={59},
  number={236},
  pages={433--460},
  year={1950}
}

@book{wiener1948cybernetics,
  title={Cybernetics: Or Control and Communication in the Animal and the Machine},
  author={Wiener, Norbert},
  year={1948},
  publisher={MIT Press}
}

@book{parker2003blink,
  title     = {In the Blink of an Eye: How Vision Sparked the Big Bang of Evolution},
  author    = {Parker, Andrew},
  publisher = {Free Press},
  year      = {2003}
}

@article{zhao2013complexity,
  title   = {Complexity and Diversity of Eyes in Early Cambrian Ecosystems},
  author  = {Zhao, Fangchen and Bottjer, David J. and Hu, Shixue and Yin, Zongjun and Zhu, Maoyan},
  journal = {Scientific Reports},
  volume  = {3},
  pages   = {2751},
  year    = {2013},
  doi     = {10.1038/srep02751}
}

@book{fitch2010evolution,
  title     = {The Evolution of Language},
  author    = {Fitch, W. Tecumseh},
  publisher = {Cambridge University Press},
  year      = {2010}
}

@book{daniels1996world,
  title     = {The World's Writing Systems},
  editor    = {Daniels, Peter T. and Bright, William},
  publisher = {Oxford University Press},
  year      = {1996}
}

@book{baugh2013history,
  title     = {A History of the English Language},
  author    = {Baugh, Albert C. and Cable, Thomas},
  edition   = {6},
  publisher = {Routledge},
  year      = {2013}
}

@article{bengio2003neural,
  title   = {A Neural Probabilistic Language Model},
  author  = {Bengio, Yoshua and Ducharme, R{\'e}jean and Vincent, Pascal and Janvin, Christian},
  journal = {Journal of Machine Learning Research},
  volume  = {3},
  pages   = {1137--1155},
  year    = {2003}
}

@article{steervit,
      title={Steerable Visual Representations}, 
      author={Jona Ruthardt and Manu Gaur and Deva Ramanan and Makarand Tapaswi and Yuki M. Asano},
      journal={ECCV},
      year={2026}
}

@misc{mikolov2013efficient,
  title        = {Efficient Estimation of Word Representations in Vector Space},
  author       = {Mikolov, Tomas and Chen, Kai and Corrado, Greg and Dean, Jeffrey},
  year         = {2013},
  eprint       = {1301.3781},
  archivePrefix= {arXiv},
  primaryClass = {cs.CL}
}

@inproceedings{sutskever2014sequence,
  title     = {Sequence to Sequence Learning with Neural Networks},
  author    = {Sutskever, Ilya and Vinyals, Oriol and Le, Quoc V.},
  booktitle = {Advances in Neural Information Processing Systems},
  volume    = {27},
  year      = {2014}
}

@inproceedings{bahdanau2015neural,
  title     = {Neural Machine Translation by Jointly Learning to Align and Translate},
  author    = {Bahdanau, Dzmitry and Cho, Kyunghyun and Bengio, Yoshua},
  booktitle = {International Conference on Learning Representations},
  year      = {2015}
}

@inproceedings{vaswani2017attention,
  title     = {Attention Is All You Need},
  author    = {Vaswani, Ashish and Shazeer, Noam and Parmar, Niki and Uszkoreit, Jakob and Jones, Llion and Gomez, Aidan N. and Kaiser, Lukasz and Polosukhin, Illia},
  booktitle = {Advances in Neural Information Processing Systems},
  volume    = {30},
  year      = {2017}
}

@inproceedings{devlin2019bert,
  title     = {{BERT}: Pre-training of Deep Bidirectional Transformers for Language Understanding},
  author    = {Devlin, Jacob and Chang, Ming-Wei and Lee, Kenton and Toutanova, Kristina},
  booktitle = {Proceedings of the 2019 Conference of the North American Chapter of the Association for Computational Linguistics: Human Language Technologies},
  pages     = {4171--4186},
  year      = {2019}
}

@misc{radford2018improving,
  title        = {Improving Language Understanding by Generative Pre-Training},
  author       = {Radford, Alec and Narasimhan, Karthik and Salimans, Tim and Sutskever, Ilya},
  year         = {2018},
  howpublished = {OpenAI Technical Report}
}

@inproceedings{brown2020language,
  title     = {Language Models are Few-Shot Learners},
  author    = {Brown, Tom B. and Mann, Benjamin and Ryder, Nick and Subbiah, Melanie and Kaplan, Jared and Dhariwal, Prafulla and Neelakantan, Arvind and Shyam, Pranav and Sastry, Girish and Askell, Amanda and Agarwal, Sandhini and Herbert-Voss, Ariel and Krueger, Gretchen and Henighan, Tom and Child, Rewon and Ramesh, Aditya and Ziegler, Daniel M. and Wu, Jeffrey and Winter, Clemens and Hesse, Christopher and Chen, Mark and Sigler, Eric and Litwin, Mateusz and Gray, Scott and Chess, Benjamin and Clark, Jack and Berner, Christopher and McCandlish, Sam and Radford, Alec and Sutskever, Ilya and Amodei, Dario},
  booktitle = {Advances in Neural Information Processing Systems},
  volume    = {33},
  pages     = {1877--1901},
  year      = {2020}
}

@article{wei2022emergent,
  title   = {Emergent Abilities of Large Language Models},
  author  = {Wei, Jason and Tay, Yi and Bommasani, Rishi and Raffel, Colin and Zoph, Barret and Borgeaud, Sebastian and Yogatama, Dani and Bosma, Maarten and Zhou, Denny and Metzler, Donald and Chi, Ed H. and Hashimoto, Tatsunori and Vinyals, Oriol and Liang, Percy and Dean, Jeff and Fedus, William},
  journal = {Transactions on Machine Learning Research},
  year    = {2022}
}

@misc{sutton2019bitter,
  title        = {The Bitter Lesson},
  author       = {Sutton, Richard S.},
  year         = {2019},
  howpublished = {\url{http://www.incompleteideas.net/IncIdeas/BitterLesson.html}}
}

@inproceedings{krizhevsky2012imagenet,
  title     = {ImageNet Classification with Deep Convolutional Neural Networks},
  author    = {Krizhevsky, Alex and Sutskever, Ilya and Hinton, Geoffrey E.},
  booktitle = {Advances in Neural Information Processing Systems},
  volume    = {25},
  year      = {2012}
}

@article{russakovsky2015imagenet,
  title   = {ImageNet Large Scale Visual Recognition Challenge},
  author  = {Russakovsky, Olga and Deng, Jia and Su, Hao and Krause, Jonathan and Satheesh, Sanjeev and Ma, Sean and Huang, Zhiheng and Karpathy, Andrej and Khosla, Aditya and Bernstein, Michael and Berg, Alexander C. and Fei-Fei, Li},
  journal = {International Journal of Computer Vision},
  volume  = {115},
  number  = {3},
  pages   = {211--252},
  year    = {2015},
  doi     = {10.1007/s11263-015-0816-y}
}

@inproceedings{radford2021learning,
  title     = {Learning Transferable Visual Models From Natural Language Supervision},
  author    = {Radford, Alec and Kim, Jong Wook and Hallacy, Chris and Ramesh, Aditya and Goh, Gabriel and Agarwal, Sandhini and Sastry, Girish and Askell, Amanda and Mishkin, Pamela and Clark, Jack and Krueger, Gretchen and Sutskever, Ilya},
  booktitle = {Proceedings of the 38th International Conference on Machine Learning},
  pages     = {8748--8763},
  year      = {2021}
}

@inproceedings{alayrac2022flamingo,
  title     = {Flamingo: A Visual Language Model for Few-Shot Learning},
  author    = {Alayrac, Jean-Baptiste and Donahue, Jeff and Luc, Pauline and Miech, Antoine and Barr, Iain and Hasson, Yana and Lenc, Karel and Mensch, Arthur and Millican, Katie and Reynolds, Malcolm and Ring, Roman and Rutherford, Eliza and Cabi, Serkan and Han, Tengda and Gong, Zhitao and Samangooei, Sina and Monteiro, Marianne and Menick, Jacob and Borgeaud, Sebastian and Brock, Andrew and Nematzadeh, Aida and Sharifzadeh, Sahand and Bińkowski, Mikołaj and Barreira, Ricardo and Vinyals, Oriol and Zisserman, Andrew and Simonyan, Karen},
  booktitle = {Advances in Neural Information Processing Systems},
  volume    = {35},
  pages     = {23716--23736},
  year      = {2022}
}

@inproceedings{li2022blip,
  title     = {{BLIP}: Bootstrapping Language-Image Pre-training for Unified Vision-Language Understanding and Generation},
  author    = {Li, Junnan and Li, Dongxu and Xiong, Caiming and Hoi, Steven},
  booktitle = {Proceedings of the 39th International Conference on Machine Learning},
  pages     = {12888--12900},
  year      = {2022}
}

@inproceedings{li2023blip2,
  title     = {{BLIP-2}: Bootstrapping Language-Image Pre-training with Frozen Image Encoders and Large Language Models},
  author    = {Li, Junnan and Li, Dongxu and Savarese, Silvio and Hoi, Steven},
  booktitle = {Proceedings of the 40th International Conference on Machine Learning},
  pages     = {19730--19742},
  year      = {2023}
}

@inproceedings{liu2023visual,
  title     = {Visual Instruction Tuning},
  author    = {Liu, Haotian and Li, Chunyuan and Wu, Qingyang and Lee, Yong Jae},
  booktitle = {Advances in Neural Information Processing Systems},
  volume    = {36},
  year      = {2023}
}

@misc{vgiworkshop2026,
  title        = {{CVPR 2026 Workshop on Visual General Intelligence: Vision Research Toward the AGI Era}},
  author       = {{VGI Workshop Organizers}},
  year         = {2026},
  howpublished = {\url{https://cvpr2026-vgi-workshop.limitlab.xyz/}},
  note         = {Accessed: 1 August, 2026}
}

@article{geirhos2020shortcut,
  title={Shortcut learning in deep neural networks},
  author={Geirhos, Robert and Jacobsen, J{\"o}rn-Henrik and Michaelis, Claudio and Zemel, Richard and Brendel, Wieland and Bethge, Matthias and Wichmann, Felix A},
  journal={Nature Machine Intelligence},
  volume={2},
  number={11},
  pages={665--673},
  year={2020},
  publisher={Nature Publishing Group UK London}
}

@article{wiedemer2025video,
  title={Video models are zero-shot learners and reasoners},
  author={Wiedemer, Thadd{\"a}us and Li, Yuxuan and Vicol, Paul and Gu, Shixiang Shane and Matarese, Nick and Swersky, Kevin and Kim, Been and Jaini, Priyank and Geirhos, Robert},
  journal={arXiv preprint arXiv:2509.20328},
  year={2025}
}

@inproceedings{geirhos2019imagenet,
  title={{ImageNet-trained CNNs are biased towards texture; increasing shape bias improves accuracy and robustness}},
  author={Geirhos, Robert and Rubisch, Patricia and Michaelis, Claudio and Bethge, Matthias and Wichmann, Felix A and Brendel, Wieland},
  booktitle={International conference on learning representations},
  year={2019}
}

@inproceedings{beery2018recognition,
  title={Recognition in terra incognita},
  author={Beery, Sara and Van Horn, Grant and Perona, Pietro},
  booktitle={Proceedings of the European conference on computer vision (ECCV)},
  pages={456--473},
  year={2018}
}

@inproceedings{zhang2026image,
  title={Are Image-to-Video Models Good Zero-Shot Image Editors?},
  author={Zhang, Zechuan and Chen, Zhenyuan and Yang, Zongxin and Yang, Yi},
  booktitle={Proceedings of the IEEE/CVF Conference on Computer Vision and Pattern Recognition},
  pages={2090--2103},
  year={2026}
}

@inproceedings{tong2026thinking,
  title={Thinking with video: Video generation as a promising multimodal reasoning paradigm},
  author={Tong, Jingqi and Mou, Yurong and Li, Hangcheng and Li, Mingzhe and Yang, Yongzhuo and Zhang, Ming and Chen, Qiguang and Liang, Tianyi and Hu, Xiaomeng and Zheng, Yining and others},
  booktitle={Proceedings of the IEEE/CVF Conference on Computer Vision and Pattern Recognition},
  pages={41121--41129},
  year={2026}
}

@inproceedings{guo2026video,
  title={Are video models ready as zero-shot reasoners? an empirical study with the mme-cof benchmark},
  author={Guo, Ziyu and Chen, Xinyan and Zhang, Renrui and An, Ruichuan and Qi, Yu and Jiang, Dongzhi and Li, Xiangtai and Zhang, Manyuan and Li, Hongsheng and Heng, Pheng-Ann},
  booktitle={Proceedings of the IEEE/CVF Conference on Computer Vision and Pattern Recognition},
  pages={9175--9184},
  year={2026}
}

@article{wang2026very,
  title={A very big video reasoning suite},
  author={Wang, Maijunxian and Wang, Ruisi and Lin, Juyi and Ji, Ran and Wiedemer, Thadd{\"a}us and Gao, Qingying and Luo, Dezhi and Qian, Yaoyao and Huang, Lianyu and Hong, Zelong and others},
  journal={arXiv preprint arXiv:2602.20159},
  year={2026}
}

@article{zhu2026video,
  title={Video Models Can Reason with Verifiable Rewards},
  author={Zhu, Tinghui and Zhang, Sheng and Huang, James Y and Song, Selena and Wen, Xiaofei and Li, Yuankai and Poon, Hoifung and Chen, Muhao},
  journal={arXiv preprint arXiv:2605.15458},
  year={2026}
}

@article{kim2026collabvr,
  title={CollabVR: Collaborative Video Reasoning with Vision-Language and Video Generation Models},
  author={Kim, Joowon and Shin, Seungho and Park, Joonhyung and Yang, Eunho},
  journal={arXiv preprint arXiv:2605.08735},
  year={2026}
}

@article{newman2026video,
  title={Video Models Reason Early: Exploiting Plan Commitment for Maze Solving},
  author={Newman, Kaleb and Zhu, Tyler and Russakovsky, Olga},
  journal={arXiv preprint arXiv:2603.30043},
  year={2026}
}

@article{liu2025can,
  title={Can World Simulators Reason? Gen-ViRe: A Generative Visual Reasoning Benchmark},
  author={Liu, Xinxin and Xu, Zhaopan and Li, Ming and Wang, Kai and Lee, Yong Jae and Shang, Yuzhang},
  journal={arXiv preprint arXiv:2511.13853},
  year={2025}
}

@article{yang2024video,
  title={Video as the new language for real-world decision making},
  author={Yang, Sherry and Walker, Jacob and Parker-Holder, Jack and Du, Yilun and Bruce, Jake and Barreto, Andre and Abbeel, Pieter and Schuurmans, Dale},
  journal={arXiv preprint arXiv:2402.17139},
  year={2024}
}

@inproceedings{yang2024depth,
  title={Depth anything: Unleashing the power of large-scale unlabeled data},
  author={Yang, Lihe and Kang, Bingyi and Huang, Zilong and Xu, Xiaogang and Feng, Jiashi and Zhao, Hengshuang},
  booktitle={Proceedings of the IEEE/CVF conference on computer vision and pattern recognition},
  pages={10371--10381},
  year={2024}
}

@inproceedings{kirillov2023segment,
  title={Segment anything},
  author={Kirillov, Alexander and Mintun, Eric and Ravi, Nikhila and Mao, Hanzi and Rolland, Chloe and Gustafson, Laura and Xiao, Tete and Whitehead, Spencer and Berg, Alexander C and Lo, Wan-Yen and others},
  booktitle={Proceedings of the IEEE/CVF international conference on computer vision},
  pages={4015--4026},
  year={2023}
}

@book{marr1982vision,
  title     = {Vision: A Computational Investigation into the Human Representation and Processing of Visual Information},
  author    = {Marr, David},
  year      = {1982},
  publisher = {W. H. Freeman},
  address   = {San Francisco}
}

@book{gibson1979ecological,
  title     = {The Ecological Approach to Visual Perception},
  author    = {Gibson, James J.},
  year      = {1979},
  publisher = {Houghton Mifflin},
  address   = {Boston}
}

@book{yarbus1967eye,
  title     = {Eye Movements and Vision},
  author    = {Yarbus, Alfred L.},
  year      = {1967},
  publisher = {Plenum Press},
  address   = {New York}
}

@article{ullman1984visual,
  title   = {Visual Routines},
  author  = {Ullman, Shimon},
  journal = {Cognition},
  volume  = {18},
  number  = {1--3},
  pages   = {97--159},
  year    = {1984}
}

@article{ballard1991animate,
  title   = {Animate Vision},
  author  = {Ballard, Dana H.},
  journal = {Artificial Intelligence},
  volume  = {48},
  number  = {1},
  pages   = {57--86},
  year    = {1991}
}

@article{liu2024decade,
  title   = {A Decade's Battle on Dataset Bias: Are We There Yet?},
  author  = {Liu, Zhuang and He, Kaiming},
  journal = {arXiv preprint arXiv:2403.08632},
  year    = {2024}
}

@inproceedings{zeng2024understanding,
  title     = {Understanding Bias in Large-Scale Visual Datasets},
  author    = {Zeng, Boya and Yin, Yida and Liu, Zhuang},
  booktitle = {Advances in Neural Information Processing Systems},
  year      = {2024}
}

@inproceedings{driess2023palme,
  title     = {PaLM-E: An Embodied Multimodal Language Model},
  author    = {Driess, Danny and Xia, Fei and Sajjadi, Mehdi S. M. and Lynch, Corey and Chowdhery, Aakanksha and Ichter, Brian and Wahid, Ayzaan and Tompson, Jonathan and Vuong, Quan and Yu, Tianhe and Huang, Wenlong and Chebotar, Yevgen and Sermanet, Pierre and Duckworth, Daniel and Levine, Sergey and Vanhoucke, Vincent and Hausman, Karol and Toussaint, Marc and Greff, Klaus and Zeng, Andy and Mordatch, Igor and Florence, Pete},
  booktitle = {Proceedings of the 40th International Conference on Machine Learning},
  series    = {Proceedings of Machine Learning Research},
  volume    = {202},
  year      = {2023}
}

@Article{Davison:ARXIV2018,
  title={{FutureMapping}: The Computational Structure of {Spatial AI} Systems},
  author={Davison, A. J.},
  journal={arXiv preprint arXiv:1803.11288},
  year={2018}
}

@Article{Davison:Ortiz:ARXIV2019,
  title={{FutureMapping 2: Gaussian Belief Propagation for Spatial AI}},
  author={Davison, A. J. and Ortiz, J.},
  journal={arXiv preprint arXiv:1910.14139},
  year={2019}
}

@inproceedings{du2021curious,
  title={Curious representation learning for embodied intelligence},
  author={Du, Yilun and Gan, Chuang and Isola, Phillip},
  booktitle={Proceedings of the IEEE/CVF International Conference on Computer Vision},
  pages={10408--10417},
  year={2021}
}

@article{wang2025compositional,
  title={Compositional scene understanding through inverse generative modeling},
  author={Wang, Yanbo and Dauwels, Justin and Du, Yilun},
  journal={arXiv preprint arXiv:2505.21780},
  year={2025}
}

@article{li2026structured,
  title={Structured 4D Latent Predictive Model for Robot Planning},
  author={Li, Zhiyi and Wu, Peilin and Han, Xiaoshen and Cai, Ruojin and Du, Yilun},
  journal={arXiv preprint arXiv:2607.01166},
  year={2026}
}

@article{du2024compositional,
  title={Compositional generative modeling: A single model is not all you need},
  author={Du, Yilun and Kaelbling, Leslie},
  journal={arXiv preprint arXiv:2402.01103},
  year={2024}
}

@article{zhou2024robodreamer,
  title={Robodreamer: Learning compositional world models for robot imagination},
  author={Zhou, Siyuan and Du, Yilun and Chen, Jiaben and Li, Yandong and Yeung, Dit-Yan and Gan, Chuang},
  journal={arXiv preprint arXiv:2404.12377},
  year={2024}
}

@article{yuille2006vision,
  title={Vision as Bayesian inference: analysis by synthesis?},
  author={Yuille, Alan and Kersten, Daniel},
  journal={Trends in cognitive sciences},
  volume={10},
  number={7},
  pages={301--308},
  year={2006},
  publisher={Elsevier}
}

@inproceedings{kulkarni2015picture,
  title={Picture: A probabilistic programming language for scene perception},
  author={Kulkarni, Tejas D and Kohli, Pushmeet and Tenenbaum, Joshua B and Mansinghka, Vikash},
  booktitle={Proceedings of the ieee conference on computer vision and pattern recognition},
  pages={4390--4399},
  year={2015}
}

@article{held1963movement,
  title={Movement-Produced Stimulation in the Development of Visually Guided Behavior},
  author={Held, Richard and Hein, Alan},
  journal={Journal of Comparative and Physiological Psychology},
  volume={56},
  number={5},
  pages={872--876},
  year={1963},
  publisher={American Psychological Association},
  doi={10.1037/h0040546}
}

@inproceedings{luo2026silvr,
  title     = {Self-Improving Loops for Visual Robotic Planning},
  author    = {Luo, Calvin and Zeng, Zilai and Jia, Mingxi and
               Du, Yilun and Sun, Chen},
  booktitle = {International Conference on Learning Representations},
  year      = {2026}
}

@article{liu2026wav,
  title   = {World Action Verifier: Self-Improving World Models via
             Forward-Inverse Asymmetry},
  author  = {Liu, Yuejiang and Feng, Fan and Kong, Lingjing and
             Lu, Weifeng and Tang, Jinzhou and Zhang, Kun and
             Murphy, Kevin and Finn, Chelsea and Du, Yilun},
  journal = {arXiv preprint arXiv:2604.01985},
  year    = {2026}
}

@incollection{mccloskey1989catastrophic,
  title     = {Catastrophic Interference in Connectionist Networks:
               The Sequential Learning Problem},
  author    = {McCloskey, Michael and Cohen, Neal J.},
  booktitle = {Psychology of Learning and Motivation},
  volume    = {24},
  pages     = {109--165},
  publisher = {Academic Press},
  year      = {1989},
  doi       = {10.1016/S0079-7421(08)60536-8}
}

@article{du2023learning,
  title={Learning universal policies via text-guided video generation},
  author={Du, Yilun and Yang, Sherry and Dai, Bo and Dai, Hanjun and Nachum, Ofir and Tenenbaum, Josh and Schuurmans, Dale and Abbeel, Pieter},
  journal={Advances in neural information processing systems},
  volume={36},
  pages={9156--9172},
  year={2023}
}

@article{chen2025large,
  title={Large video planner enables generalizable robot control},
  author={Chen, Boyuan and Zhang, Tianyuan and Geng, Haoran and Zhang, Caiyi and Li, Peihao and Song, Kiwhan and Freeman, William T and Malik, Jitendra and Abbeel, Pieter and Tedrake, Russ and others},
  journal={arXiv preprint arXiv:2512.15840},
  year={2025}
}

@inproceedings{du2022learning,
  title={Learning object-based state estimators for household robots},
  author={Du, Yilun and Lozano-Perez, Tomas and Kaelbling, Leslie Pack},
  booktitle={2022 IEEE/RSJ International Conference on Intelligent Robots and Systems (IROS)},
  pages={12558--12565},
  year={2022},
  organization={IEEE}
}

@article{fu2023neuse,
  title={Neuse: Neural se (3)-equivariant embedding for consistent spatial understanding with objects},
  author={Fu, Jiahui and Du, Yilun and Singh, Kurran and Tenenbaum, Joshua B and Leonard, John J},
  journal={arXiv preprint arXiv:2303.07308},
  year={2023}
}

@inproceedings{yang20253d,
  title={3D-mem: 3D scene memory for embodied exploration and reasoning},
  author={Yang, Yuncong and Yang, Han and Zhou, Jiachen and Chen, Peihao and Zhang, Hongxin and Du, Yilun and Gan, Chuang},
  booktitle={Proceedings of the Computer Vision and Pattern Recognition Conference},
  pages={17294--17303},
  year={2025}
}

@article{zhen2025tesseract,
  title={Tesseract: learning 4d embodied world models},
  author={Zhen, Haoyu and Sun, Qiao and Zhang, Hongxin and Li, Junyan and Zhou, Siyuan and Du, Yilun and Gan, Chuang},
  journal={arXiv preprint arXiv:2504.20995},
  year={2025}
}

@TechReport{Pearl:TechReport2017,
  author = {J. Pearl},
  title = {Theoretical Impediments to Machine Learning,
With Seven Sparks from the Causal Revolution},
  institution = {University of California, Los Angeles},
  year = {2017},
  note = {Technical Report R-275}
}

@inproceedings{murai2025,
  title={{MASt3R-SLAM}: Real-Time Dense {SLAM} with {3D} Reconstruction Priors},
  author={Murai, R.  and Dexheimer, E. and Davison, A. J.},
  booktitle=CVPR,
  year={2025}
}

@article{psu2024,
  title={Physical scene understanding},
  author={Wu, Jiajun},
  journal={AI Magazine},
  volume={45},
  number={1},
  pages={121--129},
  year={2024}
}

@phdthesis{phdthesis2019,
  title={Learning to See the Physical World},
  author={Wu, Jiajun},
  school={Massachusetts Institute of Technology},
  year={2019}
}

@inproceedings{magicpony2023,
  title={{MagicPony}: Learning Articulated {3D} Animals in the Wild},
  author={Wu, Shangzhe and Li, Ruining and Jakab, Tomas and Rupprecht, Christian and Vedaldi, Andrea},
  booktitle={CVPR},
  year={2023}
}

@inproceedings{3dfauna2024,
  title={Learning the {3D} Fauna of the Web},
  author={Li, Zizhang and Litvak, Dor and Li, Ruining and Zhang, Yunzhi and Jakab, Tomas and Rupprecht, Christian and Wu, Shangzhe and Vedaldi, Andrea and Wu, Jiajun},
  booktitle={CVPR},
  year={2024}
}

@inproceedings{objectintrinsics2023,
  title={Seeing a Rose in Five Thousand Ways},
  author={Zhang, Yunzhi and Wu, Shangzhe and Snavely, Noah and Wu, Jiajun},
  booktitle={CVPR},
  year={2023}
}

@inproceedings{rose4d2025,
  title={Birth and Death of a Rose},
  author={Geng, Chen and Zhang, Yunzhi and Wu, Shangzhe and Wu, Jiajun},
  booktitle={CVPR},
  year={2025}
}

@inproceedings{anymate2025,
  title={{Anymate}: A Dataset and Baselines for Learning {3D} Object Rigging},
  author={Deng, Yufan and Zhang, Yuhao and Geng, Chen and Wu, Shangzhe and Wu, Jiajun},
  booktitle={SIGGRAPH},
  year={2025}
}

@inproceedings{particulate2026,
  title={{Particulate}: Feed-Forward {3D} Object Articulation},
  author={Li, Ruining and Yao, Yuxin and Zheng, Chuanxia and Rupprecht, Christian and Lasenby, Joan and Wu, Shangzhe and Vedaldi, Andrea},
  booktitle={CVPR},
  year={2026}
}

@inproceedings{galileo2015,
  title={{Galileo}: Perceiving Physical Object Properties by Integrating a Physics Engine with Deep Learning},
  author={Wu, Jiajun and Yildirim, Ilker and Lim, Joseph J and Freeman, William T and Tenenbaum, Joshua B},
  booktitle={NeurIPS},
  year={2015}
}

@inproceedings{vda2017,
  title={Learning to See Physics via Visual De-animation},
  author={Wu, Jiajun and Lu, Erika and Kohli, Pushmeet and Freeman, William T and Tenenbaum, Joshua B},
  booktitle={NeurIPS},
  year={2017}
}

@article{dynamicssurvey2025,
  title={A Review of Learning-Based Dynamics Models for Robotic Manipulation},
  author={Ai, Bo and Tian, Stephen and Shi, Haochen and Wang, Yixuan and Pfaff, Tobias and Tan, Cheston and Christensen, Henrik I and Su, Hao and Wu, Jiajun and Li, Yunzhu},
  journal={Science Robotics},
  volume={10},
  number={105},
  year={2025}
}

@inproceedings{worldscore2025,
  title={{WorldScore}: A Unified Evaluation Benchmark for World Generation},
  author={Duan, Haoyi and Yu, Hong-Xing and Chen, Sirui and Fei-Fei, Li and Wu, Jiajun},
  booktitle={ICCV},
  year={2025}
}

@inproceedings{enact2026,
  title={{ENACT}: Evaluating Embodied Cognition with World Modeling of Egocentric Interaction},
  author={Wang, Qineng and Huang, Wenlong and Zhou, Yu and Yin, Hang and Bao, Tianwei and Lyu, Jianwen and Liu, Weiyu and Zhang, Ruohan and Wu, Jiajun and Fei-Fei, Li and Li, Manling},
  booktitle={ICLR},
  year={2026}
}

@inproceedings{hybridworldrep2026,
  title={Discovering Hybrid World Representations with Co-Evolving Foundation Models},
  author={Wu, Jiajun and Zhang, Yunzhi and Yu, Hong-Xing and Hsu, Joy and Mao, Jiayuan},
  booktitle={AAAI Conference on Artificial Intelligence, Emerging Trends in AI},
  year={2026}
}

@inproceedings{scenelang2025,
  title={The Scene Language: Representing Scenes with Programs, Words, and Embeddings},
  author={Zhang, Yunzhi and Li, Zizhang and Zhou, Matt and Wu, Shangzhe and Wu, Jiajun},
  booktitle={CVPR},
  year={2025}
}

@inproceedings{nscl2019,
  title={The Neuro-Symbolic Concept Learner: Interpreting Scenes, Words, and Sentences from Natural Supervision},
  author={Mao, Jiayuan and Gan, Chuang and Kohli, Pushmeet and Tenenbaum, Joshua B and Wu, Jiajun},
  booktitle={ICLR},
  year={2019}
}

@inproceedings{left2023,
  title={What's Left? Concept Grounding with Logic-Enhanced Foundation Models},
  author={Hsu, Joy and Mao, Jiayuan and Tenenbaum, Joshua B and Wu, Jiajun},
  booktitle={NeurIPS},
  year={2023}
}

@article{nsconcept2026,
  title={Building Intelligent Agents with Neuro-Symbolic Concepts},
  author={Mao, Jiayuan and Tenenbaum, Joshua B and Wu, Jiajun},
  journal={Communications of the ACM},
  volume={69},
  number={2},
  year={2026}
}

@article{articraft2026,
  title={{Articraft}: An Agentic System for Scalable Articulated {3D} Asset Generation},
  author={Zhou, Matt and Li, Ruining and Lyu, Xiaoyang and Song, Zhaomou and Huang, Zhening and Zheng, Chuanxia and Rupprecht, Christian and Vedaldi, Andrea and Wu, Shangzhe},
  journal={arXiv preprint arXiv:2605.15187},
  year={2026}
}

@inproceedings{3dgeneralist2026,
  title={{3D-Generalist}: Vision-Language-Action Models for Crafting {3D} Worlds},
  author={Sun, Fan-Yun and Wu, Shengguang and Jacobsen, Christian and Yim, Thomas and Zou, Haoming and Zook, Alex and Li, Shangru and Chou, Yu-Hsin and Can, Ethem and Wu, Xunlei and Eppner, Clemens and Blukis, Valts and Tremblay, Jonathan and Wu, Jiajun and Birchfield, Stan and Haber, Nick},
  booktitle={3DV},
  year={2026}
}

@inproceedings{capx2026,
  title={{CaP-X}: A Framework for Benchmarking and Improving Coding Agents for Robot Manipulation},
  author={Fu, Letian and Yu, Justin and El-Refai, Karim and Kou, Ethan and Xue, Haoru and Huang, Huang and Xiao, Wenli and Wang, Guanzhi and Niu, Dantong and Fei-Fei, Li and Shi, Guanya and Wu, Jiajun and Sastry, Shankar and Zhu, Yuke and Goldberg, Ken and Fan, Linxi},
  booktitle={ICML},
  year={2026}
}

@article{li2026instructparticulate,
  title   = {{Instruct-Particulate}: Scaling Feed-Forward 3D Object Articulation with Kinematic Control},
  author  = {Li, Ruining and Yao, Yuxin and Zhou, Matt and Zheng, Chuanxia and Rupprecht, Christian and Lasenby, Joan and Wu, Shangzhe and Vedaldi, Andrea},
  journal = {arXiv preprint arXiv:2606.14699},
  year    = {2026}
}

@InProceedings{geng2026neurok,
  title     = {{NeuROK}: Generative 4D Neural Object Kinematics},
  author    = {Chen Geng and Guangzhao He and Yue Gao and Yunzhi Zhang and Shangzhe Wu and Jiajun Wu},
  booktitle = {CVPR},
  year      = {2026}
}

@InProceedings{lyu2026choreograph,
  title     = {Choreographing a World of Dynamic Objects},
  author    = {Yanzhe Lyu and Chen Geng and Karthik Dharmarajan and Yunzhi Zhang and Hadi Alzayer and Shangzhe Wu and Jiajun Wu},
  booktitle = {CVPR},
  year      = {2026}
}

@article{berv2026rates,
  title={Rates of passerine body plan evolution in time and space},
  author={Berv, Jacob S and Probst, Charlotte M and Claramunt, Santiago and Shipley, J Ryan and Friedman, Matt and Smith, Stephen A and Fouhey, David F and Weeks, Brian C},
  journal={Nature Ecology \& Evolution},
  pages={1--15},
  year={2026},
  publisher={Nature Publishing Group UK London}
}

@article{weeks2025skeletal,
  title={Skeletal trait measurements for thousands of bird species},
  author={Weeks, Brian C and Zhou, Zhizhuo and Probst, Charlotte M and Berv, Jacob S and O’Brien, Bruce and Benz, Brett W and Skeen, Heather R and Ziebell, Mark and Bodt, Louise and Fouhey, David F},
  journal={Scientific Data},
  volume={12},
  number={1},
  pages={884},
  year={2025},
  publisher={Nature Publishing Group UK London}
}

@article{wang2024supersynthia,
  title={SuperSynthIA: Physics-ready Full-disk Vector Magnetograms from HMI, Hinode, and Machine Learning},
  author={Wang, Ruoyu and Fouhey, David F and Higgins, Richard EL and Antiochos, Spiro K and Barnes, Graham and Hoeksema, J Todd and Liu, Yang and Schuck, Peter W and Gombosi, Tamas I},
  journal={The Astrophysical Journal},
  volume={970},
  number={2},
  pages={168},
  year={2024},
  publisher={The American Astronomical Society}
}

@article{cheung2015thermal,
  title={Thermal diagnostics with the atmospheric imaging assembly on board the solar dynamics observatory: a validated method for differential emission measure inversions},
  author={Cheung, Mark CM and Boerner, P and Schrijver, CJ and Testa, P and Chen, F and Peter, H and Malanushenko, A},
  journal={The Astrophysical Journal},
  volume={807},
  number={2},
  pages={143},
  year={2015},
  publisher={The American Astronomical Society}
}

@article{hathaway2015solar,
  title={The solar cycle},
  author={Hathaway, David H},
  journal={Living reviews in solar physics},
  volume={12},
  number={1},
  pages={4},
  year={2015},
  publisher={Springer}
}

@article{brookes1976observation,
  title={Observation of free oscillations of the Sun},
  author={Brookes, JR and Isaak, GR and Van der Raay, HB},
  journal={Nature},
  volume={259},
  number={5539},
  pages={92--95},
  year={1976},
  publisher={Nature Publishing Group UK London}
}

@article{hoeksema2014helioseismic,
  title={The Helioseismic and Magnetic Imager (HMI) vector magnetic field pipeline: Overview and performance},
  author={Hoeksema, J Todd and Liu, Yang and Hayashi, Keiji and Sun, Xudong and Schou, Jesper and Couvidat, Sebastien and Norton, Aimee and Bobra, Monica and Centeno, Rebecca and Leka, KD and others},
  journal={Solar Physics},
  volume={289},
  number={9},
  pages={3483--3530},
  year={2014},
  publisher={Springer}
}

@article{higgins2021fast,
  title={Fast and accurate emulation of the SDO/HMI Stokes inversion with uncertainty quantification},
  author={Higgins, Richard EL and Fouhey, David F and Zhang, Dichang and Antiochos, Spiro K and Barnes, Graham and Hoeksema, J Todd and Leka, KD and Liu, Yang and Schuck, Peter W and Gombosi, Tamas I},
  journal={The Astrophysical Journal},
  volume={911},
  number={2},
  pages={130},
  year={2021},
  publisher={The American Astronomical Society}
}

@article{hersbach2020era5,
  title={The ERA5 global reanalysis},
  author={Hersbach, Hans and Bell, Bill and Berrisford, Paul and Hirahara, Shoji and Hor{\'a}nyi, Andr{\'a}s and Mu{\~n}oz-Sabater, Joaqu{\'\i}n and Nicolas, Julien and Peubey, Carole and Radu, Raluca and Schepers, Dinand and others},
  journal={Quarterly journal of the royal meteorological society},
  volume={146},
  number={730},
  pages={1999--2049},
  year={2020},
  publisher={Wiley Online Library}
}

@article{weeks2020shared,
  title={Shared morphological consequences of global warming in North American migratory birds},
  author={Weeks, Brian C and Willard, David E and Zimova, Marketa and Ellis, Aspen A and Witynski, Max L and Hennen, Mary and Winger, Benjamin M},
  journal={Ecology Letters},
  volume={23},
  number={2},
  pages={316--325},
  year={2020},
  publisher={Wiley Online Library}
}
}

\end{document}